\documentclass{article}

\usepackage[numbers]{natbib}

\usepackage[preprint]{arxiv_2021}

\usepackage[utf8]{inputenc} 
\usepackage[T1]{fontenc}    
\usepackage{hyperref}       
\usepackage{url}            
\usepackage{booktabs}       
\usepackage{amsfonts}       
\usepackage{nicefrac}       
\usepackage{microtype}      
\usepackage{xcolor}         
\usepackage{amssymb}
\usepackage{multirow}
\usepackage{graphicx}
\usepackage{subfigure}
\usepackage{wrapfig}
\usepackage{bbm}
\usepackage[noend]{algpseudocode}
\usepackage{algorithmicx,algorithm}

\usepackage{amsthm,amsmath,amssymb}
\usepackage{mathrsfs}
\usepackage{enumitem}

\title{Exploring Transferable and Robust Adversarial Perturbation Generation from the Perspective of Network Hierarchy}
\author{Ruikui Wang$^{1}$, Yuanfang Guo$^{1}$, Ruijie Yang$^{1}$, Yunhong Wang$^{1}$\\
	$^1$~School of Computer Science and Engineering, Beihang University\\
	Beijing, China\\
	{\tt\small \{rkwang,rjyang,andyguo,yhwang\}@buaa.edu.cn}}

\begin{document}

\maketitle

\begin{abstract}
  The transferability and robustness of adversarial examples are two practical yet important properties for black-box adversarial attacks. In this paper, we explore effective mechanisms to boost both of them from the perspective of network hierarchy, where a typical network can be hierarchically divided into output stage, intermediate stage and input stage. Since over-specialization of source model, we can hardly improve the transferability and robustness of the adversarial perturbations in the output stage. Therefore, we focus on the intermediate and input stages in this paper and propose a transferable and robust adversarial perturbation generation (TRAP) method. Specifically, we propose the dynamically guided mechanism to continuously calculate accurate directional guidances for perturbation generation in the intermediate stage. In the input stage, instead of the single-form transformation augmentations adopted in the existing methods, we leverage multi-form affine transformation augmentations to further enrich the input diversity and boost the robustness and transferability of the adversarial perturbations. Extensive experiments demonstrate that our TRAP achieves impressive transferability and high robustness against certain interferences.
\end{abstract}

\section{Introduction}
Deep Neural Networks (DNNs) have achieved remarkable success in various computer visual tasks, including image classification \cite{krizhevsky2012imagenet,simonyan2014very,he2016identity}, object tracking \cite{held2016learning,kosiorek2017hierarchical,li2019siamrpn++}, detection \cite{ren2015faster,pang2017towards}, semantic segmentation \cite{long2015fully,bucher2019zero,chaitanya2020contrastive}, etc. Unfortunately, these DNNs are easily to be interfered by some images, which contain certain artificial yet inconspicuous perturbations, and yield incorrect outputs.
These deliberately crafted perturbations are named adversarial perturbations \cite{szegedy2013intriguing}, which has attracted significant attentions of researchers in the past few years \cite{szegedy2013intriguing,goodfellow2014explaining,kurakin2016adversarial,papernot2016transferability,carlini2017towards,dong2018boosting,dong2019evading,wang2020unified}. In general, the adversarial perturbation generation (i.e., adversarial attack) methods can be classified into two categories: white-box attacks and black-box attacks. The white-box attacks usually assume that the attackers possess access to both the architecture and gradient information of the target model. On the contrary, the black-box attacks assume that the attackers possess approximately zero knowledge besides of the final predictions. In the black-box scenarios, the adversarial attack methods usually require the generated perturbations to possess high transferability \cite{liu2016delving}, which is vital to the attack success rate. Recently, many literatures have been published to focusing on investigating this intriguing property \cite{papernot2016transferability,dong2018boosting,zhou2018transferable,huang2019enhancing,xie2019improving,inkawhich2020transferable}. Besides, the property of an adversarial example, which assesses the performance drops against various interferences, is named robustness, which is also a practical concern in real world systems \cite{athalye2018synthesizing,eykholt2018robust,tsipras2018robustness,ross2018improving,song2018physical,arnab2018robustness,engstrom2019exploring}.

In this paper, we explore to generate adversarial perturbations with high transferability and robustness from the perspective of network hierarchy. Intuitively, a deep network architecture can be hierarchically divided into three major stages, i.e., output stage, intermediate stage and input stage. Most of the existing adversarial attack methods manipulate in one of these stages, each of which are relatively orthogonal to the others and can thus play a complementary role to achieve higher transferability and robustness.

Unfortunately, to our best knowledge, there is no existing approaches which can manipulate in the output stage and boost the transferability. We believe that this phenomenon is induced by the over-specialization of the DNN model in the output stage, which is a classic yet typical generalization problem in machine learning. Therefore, we will not devote our efforts in this stage also, in this paper. For the intermediate stage, some literatures \cite{sabour2015adversarial,rozsa2017lots,zhou2018transferable, inkawhich2019feature, huang2019enhancing,li2020yet,wu2020boosting,inkawhich2020transferable,lu2020enhancing} have explored to perturb the intermediate features to improve the transferability of adversarial examples. \cite{huang2019enhancing} develops a new paradigm to specifically optimize the transferability by utilizing the intermediate representation of a given adversarial example as directional guidance, which provides a reasonable proxy for generating the transferable perturbations. \cite{huang2019enhancing} inspires \cite{li2020yet} to further explore linear combinations of auxiliary results, which are produced in the baseline phase, as the directional guidance for the latter phase. However, since these directional guidances are fixed once the baseline phase is completed, the positive effects of these guidances tend to decline as the subsequent optimization steps carry out. Meanwhile, the transferability, as well as the robustness, of the adversarial examples benefits from creating diverse inputs at the input stage of network \cite{athalye2018synthesizing,lin2019nesterov,xie2019improving,zou2020improving,yang2021adversarial}. \cite{athalye2018synthesizing} generates robust adversarial examples by introducing Expectation Over Transformation (EOT) to the inputs. Unfortunately, \cite{athalye2018synthesizing} only applies single-form transformations, i.e. only one type of basic transformations at a time, to the input image, where the complete (multi-form) affine transformation cannot be constructed to improve the transferability and robustness of the adversarial examples. Note that, the expectation operation tends to bring larger computational overheads.


\begin{figure*}[tp]
	\centering
	\includegraphics[width=\linewidth,trim=0.4cm 3.1cm 0.4cm 2.6cm,clip]{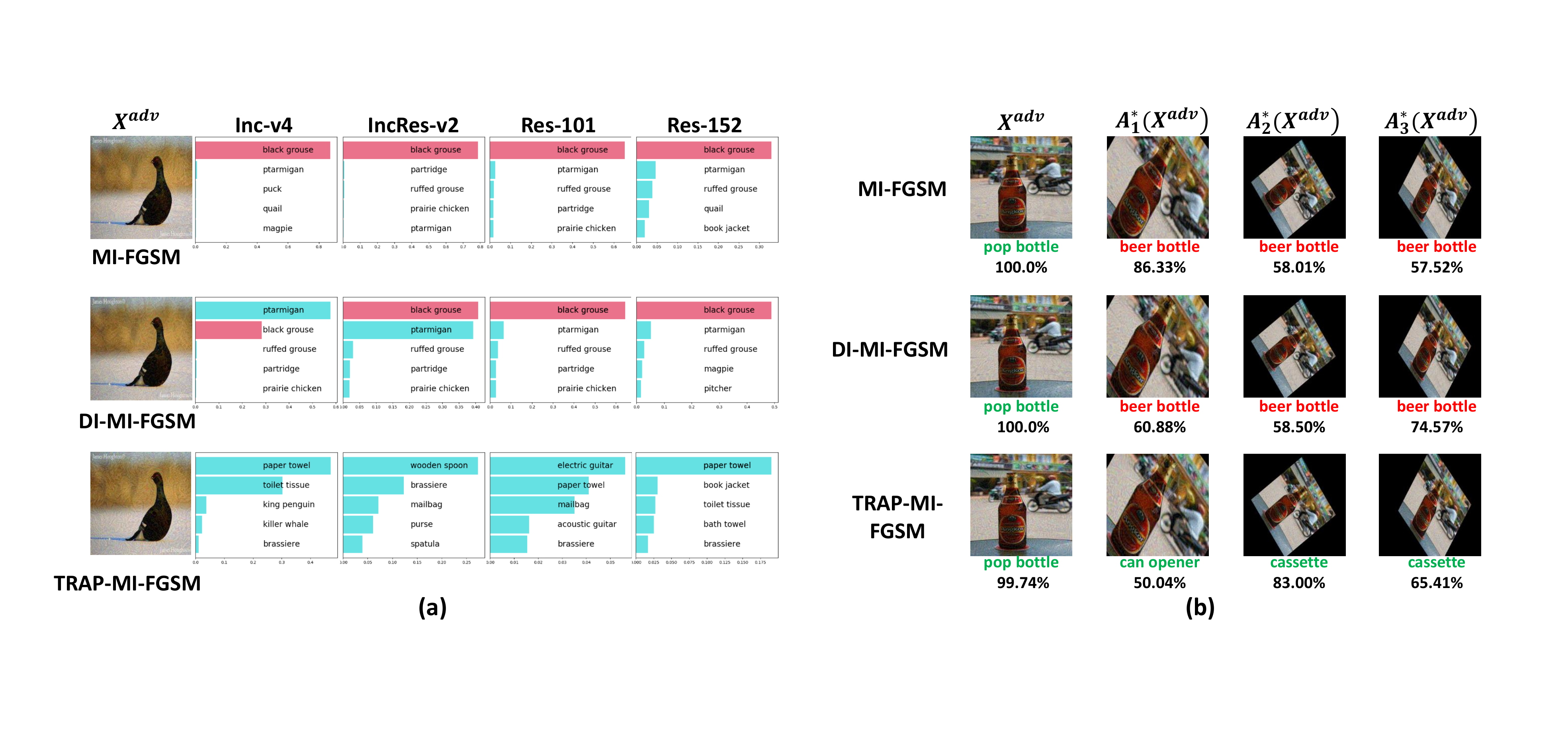}
	\caption{Demonstration of transferability and robustness of adversarial examples. All the adversarial examples are crafted on Inception-v3 with $\epsilon=16$. (a) The comparison of transferability of adversarial examples. The ground-truth label is highlighted in pink. (b) The comparison of robustness of adversarial examples. $X^{adv}$ denotes the original adversarial examples and $A^*_i(\cdot)$ denotes different multi-form transformations. Note that the descriptions under each image indicate its predicted category by Inception-v3 and corresponding confidence. We highlight the category tags in green if the attack is successful. If the attack is failed, the category tag is highlighted in red.}
	\label{Figure_AIM}
\end{figure*}

To tackle these issues, we propose a transferable and robust adversarial perturbation generation (TRAP) method from the perspective of network hierarchy. Specifically, 
dynamically guided mechanism, which adaptively updates the directional guidance as the optimization progresses, is proposed to boost the transferability in the intermediate stage. Figure \ref{Figure_AIM}(a) gives an example of the transferability property of our method. In the input stage, we introduce a multi-form affine-transformation, which contains multiple types of basic transformations, to the adversarial examples to perform combinatorial augmentations with little computational overheads. Figure \ref{Figure_AIM}(b) reveals that our primary intention of utilizing the multi-form transformation augmentation is to boost the robustness of the perturbations to the practical variations.

Our contributions are summarized as below:
\begin{itemize}[leftmargin=*]
	\setlength{\itemsep}{0pt}
	\setlength{\parsep}{0pt}
	\setlength{\parskip}{0pt}
	\item We propose a transferable and robust adversarial perturbation generation (TRAP) method from the perspective of network hierarchy to boost the transferability and robustness of the adversarial examples simultaneously.
	\item We propose a dynamically guided mechanism in the intermediate stage of network to adaptively revise the directional guidance, as the perturbation generation process performs.
	\item We propose an affine-invariant perturbation enhancement mechanism, which improves both the transferability and robustness of the adversarial perturbations, by augmenting the input images with multi-form affine transformations, in the input stage.
\end{itemize}

\section{Related Work}
In this section, we give a brief review to the related black-box attack methods. By regarding the adversarial perturbation generation process as an optimization problem, gradient-based methods usually leverage the existing optimization algorithms to boost the transferability of adversarial examples \cite{goodfellow2014explaining,kurakin2016adversarial,kurakin2017adversarial,tramer2017ensemble,dong2018boosting,lin2019nesterov,wang2021boosting}.
 
For the intermediate stage of network, perturbing the intermediate feature space is proposed to improve the transferability of the black-box methods. Transferable Adversarial Perturbations (TAP) \cite{zhou2018transferable} maximizes the distances between the benign images and their adversarial versions at all the hidden layers. To search for perturbations with better transferability, Intermediate Level Attack (ILA) \cite{huang2019enhancing} maximizes the scalar projection of the adversarial example onto a guided direction on a specific hidden layer. Motivated by ILA \cite{huang2019enhancing}, \cite{li2020yet} takes the advantage of auxiliary examples produced by a baseline attack and yields adversarial examples with better transferability.

In the input stage of network, input augmentation can be exploited to facilitate both the robustness and transferability. EOT \cite{athalye2018synthesizing} adopts this principle and generates robust adversarial examples via single-form affine transformation augmentations. Diverse Input Method (DIM) \cite{xie2019improving} further demonstrates the effectiveness of input augmentation by applying random resizing and padding to the inputs with a fixed probability. Scale-Invariant Method (SIM) \cite{lin2019nesterov} enriches the gradient information over an ensemble of multi-scale copies of input image and generates more transferable adversarial examples.

\section{Proposed Work}
Given a clean image $X$, its ground-truth label $y^{true}$ and a substitute network with parameters $\theta$, we aim to generate an adversarial perturbation for $X$ with high transferability and robustness. Considering the relative orthogonality of different stages, naturally, we can design different mechanisms for each stage and combine them to construct a novel adversarial perturbation generation method, named transferable and robust adversarial perturbation generation (TRAP). The complete procedures of our TRAP are summarized in Algorithm \ref{algorithm1}, which contains two mechanisms, i.e., \emph{dynamically guided mechanism} and \emph{affine-invariant perturbation enhancement mechanism}. Note that the baseline attack model in our method can be any transfer-based black-box attack methods with any gradient based iterative optimizer. For convenience, we simply employ MI-FGSM \cite{dong2018boosting} as our baseline model and Momentum-SGD \cite{polyak1964some} as the optimizer in this paper.

\subsection{Dynamically Guided Mechanism (DGM)}

For the intermediate stage of network, \cite{li2020yet} has empirically demonstrated that a larger perturbation on the intermediate feature leads to a higher transferability. Then, a straightforward strategy is to explicitly maximize the distances between the benign images and their corresponding adversarial examples in feature space \cite{zhou2018transferable,wu2020boosting,inkawhich2019feature}. However,  the transferability of such straightforwardly generated examples usually becomes unsatisfactory, which is induced by overfitting the source model when the number of attack iteration increases. On the contrary, \cite{huang2019enhancing, li2020yet} leverage two phases, i.e., \emph{baseline phase} and \emph{enhancement phase}, to implicitly enlarge the intermediate feature gap between the benign and perturbed images, with respect to a directional guidance obtained in the baseline phase. Unfortunately, \cite{huang2019enhancing, li2020yet} only exploit fixed directional guidance, which is generated by the baseline phase and can only provide sufficient guidelines to the initial steps of the enhancement phase. It cannot provide accurate reference information for subsequent optimization procedures. Therefore, we propose a dynamically guided mechanism to alleviate this problem.

Specifically, we firstly perform $t_{1}$ steps of the baseline attack phase and adopt cross-entropy loss to train initial the attack model as:
\begin{equation}
	\displaystyle
	{L_1}({X^{adv}},{y^{true}};\theta ) =  - {\mathbbm{1}_{{y^{true}}}}log(softmax(c(X^{adv};\theta))),
	\label{eq-2}
\end{equation}
where $\mathbbm{1}_{y^{true}}$ represents the one-hot formed ground-truth and $c(X;\theta)$ denotes the logits output and $\theta$ stands for the network parameters. Then, an initial adversarial example $X^{adv}_{t_1}$ can be obtained. In our enhancement phase, we progressively update the directional guidance as the enhancement process continues. In practice, we initialize the directional guidance $h^{*}_{t_1}$ with the hidden output $h^{adv}_{t_1}$, which can be computed by
\begin{equation}
	\displaystyle
	h^{adv}_{t_1}=flatten(\mathscr{F}^l(X^{adv}_{t_1};\theta)),
	\label{eq-3}
\end{equation}
where $\mathscr{F}^l(\cdot;\theta)$ indicates the output function of layer $l$ of source model. Then, it can be updated in a progressive manner via
\begin{equation}
	\displaystyle
	h_{t}^* = (1 - \beta )h_{t-1}^{adv} + \beta h_{t-1}^*,t\ge{t_1},
	\label{eq0}
\end{equation}
where $t$ denotes the optimization step in the enhancement phase, $h_t^{adv}$ represents the hidden output of $X^{adv}_t$, and $\beta$ is employed to balance the tradeoff between the historical and current directional guidances. Then, the gradient optimization direction can be sought for, with the help of the above dynamically updated directional guidance. Similar to \cite{huang2019enhancing}, we not only expect the optimized direction $h^{adv}_t$ to be consistent with the dynamical directional guidance $h_t^*$, but also anticipate a larger amplitude of the perturbation in that direction. This optimization can be achieved via:
\begin{equation}
	\displaystyle
	{L_2}(X_t^{adv}, h_t^{*};\theta ) = \frac{{<h_t^* - {h^x},h_t^{adv} - {h^x}>}}{{{{\left\| {h_t^* - {h^x}} \right\|}_2}{{\left\| {h_t^{adv} - {h^x}} \right\|}_2}}} + \gamma \frac{{{{\left\| {h_t^{adv} - {h^x}} \right\|}_2}}}{{{{\left\| {h_t^* - {h^x}} \right\|}_2}}},t \ge {t_1},
	\label{eq-1}
\end{equation}
where $h^x$ and $h_t^{adv}$ stand for the hidden outputs of the original image $X$ and $X_t^{adv}$, respectively, and $\gamma$ denotes the tradeoff parameter.

\begin{algorithm}[tp]
	\caption{Transferable and Robust Adversarial Perturbation Generation (TRAP)}
	\hspace*{0.02in} {\bf Input:} 
	A benign example $X$ with ground-truth label $y^{true}$; the source model parameter $\theta$;\\
	\hspace*{0.02in} {\bf Input:}
	Perturbation budget $\epsilon$; maximum iterations $T$; iterations of baseline attack phase $t_1$; momentum factor $\mu$; the transformation probability $p$;\\
	\hspace*{0.02in} {\bf Output:} 
	Final adversarial example $X^{adv}_T$
	\begin{algorithmic}[1]
		\State $\alpha=\epsilon/t_1$; $X^{adv}_0=X$ 
		\State $g_m=0$;$h_0^*=0$
		\For{$t=0$ to $T-1$}
		\If{$t==t_1$} 
		\State $X^{adv}_t=X$; $g_m=0$ 
		\EndIf
		\If{$t<t_1$}
		\State $g={\nabla _X}{L_1}({A_1} \circ {A_2} \circ {A_3} \circ {A_4}(X_t^{adv};p),{y^{true}}; \theta)$ 
		\Else
		\State $g={\nabla _X}{L_2}({A_1} \circ {A_2} \circ {A_3} \circ {A_4}(X_t^{adv};p), h_t^*; \theta)$
		\EndIf
		\State Update $g_{t+1}=\mu \cdot g_m + \frac{g}{{{{\left\| g \right\|}_1}}}$; $g_m=g_{t+1}$
		\If{$t>=t_1$}
		\State $\alpha=\epsilon/(T-t_1)$
		\EndIf
		\State Update $X_{t + 1}^{adv}=Clip_X^\varepsilon \{ X_t^{adv} + \alpha  \cdot sign({g_{t+1}})\}$
		\If{$t>=t_1$}
		\State Update $h_{t+1}^*$ by Eq.\ref{eq0}
		\Else
		\State $h_{t + 1}^* = flatten(\mathscr{F}^l(X^{adv}_{t+1};\theta))$
		\EndIf
		\EndFor
		\State \Return $X^{adv}_T$
	\end{algorithmic}
	\label{algorithm1}
\end{algorithm}

\subsection{Affine-Invariant Perturbation Enhancement Mechanism (AIM)}
For input stage of network, we introduce a new augmentation paradigm, i.e., imposing multiple basic instantiations of affine transformations on input concurrently to enrich the diversity of input patterns, when generating the adversarial examples. Specifically, we leverage four types of basic transformations, including translation, rotation, scaling and shearing. Note that these differentiable basic transformations can be formulated in an unified mathematical expression and the calculations can benefit from GPU accelerations. Besides, the invariance against these transformations is of great significance to facilitate the robustness of adversarial examples \cite{athalye2018synthesizing}.

Let $A_{1}(\cdot)$, $A_{2}( \cdot )$, $A_{3}( \cdot )$, $A_{4}( \cdot )$ denote the translation, rotation, scaling and shearing operations respectively. In this paper, these operations are formulated in a manner of coordinate transformation. Specifically, the formulation of translation, $A_{1}( \cdot )$, can be defined as
\begin{equation}
	\displaystyle
	\left[ \begin{array}{l}
		{x'}\\
		{y'}\\
		1
	\end{array} \right] = \left[ {\begin{array}{*{20}{c}}
			1&0&{{t_x}}\\
			0&1&{{t_y}}\\
			0&0&1
	\end{array}} \right]\left[ \begin{array}{l}
		x\\
		y\\
		1
	\end{array} \right],
	\label{eq1}
\end{equation}
where $({x}$, ${y})$ denotes the coordinates of the original image pixel, $({x'}$, ${y'})$ denotes the coordinates of the transformed image pixel, and $t_x$ and $t_y$ represent the offsets of translation in the horizontal and vertical directions, respectively. Similarly, the formulation of rotation, $A_{2}( \cdot )$, can be defined as
\begin{equation}
	\displaystyle
	\left[ \begin{array}{l}
		{x'}\\
		{y'}\\
		1
	\end{array} \right] = \left[ {\begin{array}{*{20}{c}}
			{cos(\theta )}&{ - sin(\theta )}&0\\
			{sin(\theta )}&{cos(\theta )}&0\\
			0&0&1
	\end{array}} \right]\left[ \begin{array}{l}
		x\\
		y\\
		1
	\end{array} \right],
	\label{eq2}
\end{equation}
where $\theta$ represents the degree of rotation. The formulation of scaling, $A_{3}( \cdot )$, can be defined as
\begin{equation}
	\displaystyle
	\left[ \begin{array}{l}
		{x'}\\
		{y'}\\
		1
	\end{array} \right] = \left[ {\begin{array}{*{20}{c}}
			{{s_x}}&0&0\\
			0&{{s_y}}&0\\
			0&0&1
	\end{array}} \right]\left[ \begin{array}{l}
		x\\
		y\\
		1
	\end{array} \right],
	\label{eq3}
\end{equation}
where $s_x$ and $s_y$ denote the scaling factors of the horizontal and vertical directions, respectively. The formulation of shearing, $A_{4}( \cdot )$, can be defined as
\begin{equation}
	\displaystyle
	\left[ \begin{array}{l}
		{x'}\\
		{y'}\\
		1
	\end{array} \right] = \left[ {\begin{array}{*{20}{c}}
			1&{{d_x}}&0\\
			{{d_y}}&1&0\\
			0&0&1
	\end{array}} \right]\left[ \begin{array}{l}
		x\\
		y\\
		1
	\end{array} \right],
	\label{eq4}
\end{equation}
where $d_x$ and $d_y$ denote the two shearing factors.

With the above defined basic transformations in (\ref{eq1}) to (\ref{eq4}), a multi-form affine transformation can be achieved simply by multiplying these transformation matrices. In practice, we impose multi-form transformations on adversarial example generated in each step. The updating rules of our proposed AIM can be formulated in such a iterative manner:
\begin{equation}
	X_{t + 1}^{adv} = Clip_X^\varepsilon (X_t^{adv} + \alpha  \cdot sign({\nabla _X}L({A_1} \circ {A_2} \circ {A_3} \circ {A_4}(X_t^{adv};p),{y^{true}};\theta))),
\end{equation}
where $Clip_X^\varepsilon(\cdot)$ denotes the clip operation which ensures the generated perturbed image is within the $\epsilon-$ball of the benign image $X$. Note that ${\nabla _X}L(X,{y^{true}};\theta )$ represents the gradient of the final loss with respect to $X$. ${A_1} \circ {A_2} \circ {A_3} \circ {A_4}$ stands for our multi-form affine transformation operation, which can degenerate to the single-form one by fixing certain transformation factors or offsets. Similar to \cite{xie2019improving}, $p$ controls the execution probability of applying affine-transformation in each iteration. $t$ and $\alpha$ denote the number of iterations and step size, respectively.

\section{Experimental Results}
In this section, we evaluate our proposed TRAP in various experiments. Please note that we employ the prefix `DG-' to represent the utilization of our dynamically guided mechanism and `AI-' to represent the utilization of our affine-invariant perturbation enhancement mechanism, in the results.
\subsection{Experimental Settings}
\textbf{Dataset.}
By following the experimental protocols in \cite{lin2019nesterov,wang2021enhancing}, we randomly select $1000$ images, which belong to $1000$ classes, from the ILSVRC2012 validation set \cite{russakovsky2015imagenet}. For performance measurement, we adopt the commonly used attack success rate (ASR).

\textbf{Networks.}
We adopt five DNN models, i.e., Inception-v3(Inc-v3) \cite{szegedy2016rethinking}, Inception-v4(Inc-v4) \cite{szegedy2017inception}, Inception-Resnet-v2(IncRes-v2) \cite{szegedy2017inception}, ResNet-101(Res-101) \cite{he2016identity}, ResNet-152(Res-152) \cite{he2016identity}. All the models are available on Github\footnote{https://github.com/Cadene/pretrained-models.pytorch}.

\textbf{Implementation details.}
In the experiments, MI-FGSM is employed as our baseline model. For fair comparisons, all of the methods in the experiments employ Momentum-SGD as the optimizer and the momentum factor $\mu$ is set to $1.0$. The maximum perturbation for each pixel is set to be $\epsilon=16$. The number of iterations $T$ is set to 10. And step size is determined by $\alpha=\epsilon/T$. For DI-MI-FGSM \cite{xie2019improving} and AI-MI-FGSM, the transformation probability $p$ is set to $0.9$. For ILA \cite{huang2019enhancing} and DG-ILA, we set $\gamma$ to 0.8. For DG-ILA, we set $\beta$ to 0.8 and the number of iterations for the baseline phase $t_1$ to 4. All the input images are resized to $[299,299,3]$. As for affine-transformation, we determine $t_x,t_y$ by sampling from the uniform distribution $[-0.1,0.1]$ for each step. Similar, $\theta$ is sampled from $[-90,90]$, $s_x,s_y$ are sampled from $[0.5,1.5]$ and $d_x,d_y$ are sampled from $[-30,30]$.

\subsection{Evaluation of DGM}

\begin{wraptable}[24]{r}{0.5\textwidth}
	\centering
	\caption{The evaluation of our DGM. The source models we used are listed in the first column and the target models are listed in the first line.}
	\label{tab0}
	\resizebox{0.5\textwidth}{!}{
		\begin{tabular}{@{}c|c|ccccc@{}}
			\toprule
			Model                   & Attack         & Inc-v3  & Inc-v4 & IncRes-v2 & Res-101 & Res-152 \\ 
			\hline
			\multirow{6}{*}{Inc-v3} & TAP\cite{zhou2018transferable}        & \bf{99.8}\% & 81.1\% & 75.9\%   & 77.6\%  & 73.6\%  \\
			& TAP*     & 99.7\%  & 83.9\% & 78.2\%    & 77.1\%  & 74.4\%  \\
			& ILA\cite{huang2019enhancing} & 99.6\%  & 82.1\% & 78.6\%    & 76.8\%  & 73.8\%  \\
			& DG-ILA(ours) & 99.7\%  & 83.6\% & 81.2\%    & 79.4\%  & 76.6\%  \\
			& ILA++\cite{li2020yet}  & 99.7\%  & 85.0\% & 81.3\%    & 79.2\%       & 75.5\%   \\
			& DG-ILA++(ours) & 99.5\%  & \bf{85.6}\%  & \bf{82.5}\% & \bf{81.0}\% & \bf{76.8}\%  \\
			\hline
			\hline
			\multirow{6}{*}{Inc-v4} & TAP\cite{zhou2018transferable}        & 52.8\% & \bf{99.9}\% &  45.3\%  & 48.2\%  & 43.2\%  \\
			& TAP*     & 79.2\%  & 98.7\% & 70.3\%    & 76.2\%  & 73.4\%  \\
			& ILA\cite{huang2019enhancing} & 77.9\%  & 98.2\% & 71.3\%    & 74.7\%  & 70.4\%  \\
			& DG-ILA(ours) & 78.7\%  & 98.0\% & 72.8\%    & 77.7\%  & 73.1\%  \\
			& ILA++\cite{li2020yet}  & 78.9\%  & 99.0\% & 73.3\%    & 77.9\%       & 73.2\%   \\
			& DG-ILA++(ours) & \bf{80.2}\%  & 98.8\%  & \bf{74.8}\% & \bf{78.7}\% & \bf{74.9}\%  \\
			\hline
			\hline
			\multirow{6}{*}{IncRes-v2} & TAP\cite{zhou2018transferable}        & 62.8\% & 59.6\% & 96.0\%   & 63.6\%  & 56.8\%  \\
			& TAP*     & 79.1\%  & 74.9\% & 97.1\%    & 73.0\%  & 69.0\%  \\
			& ILA\cite{huang2019enhancing} & 76.5\%  & 74.1\% & 97.0\%    & 73.1\%  & 68.5\%  \\
			& DG-ILA(ours) & 82.1\%  & \bf{78.4}\% & 97.5\%    & 76.9\%  & 72.3\%  \\
			& ILA++\cite{li2020yet}  & 79.5\%  & 76.1\% & \bf{98.1}\%    & 75.6\%       & 69.9\%   \\
			& DG-ILA++(ours) & \bf{82.7}\%  & 78.1\%  & \bf{98.1}\% & \bf{78.0}\% & \bf{72.7}\%  \\
			\hline
			\hline
			\multirow{6}{*}{Res-101} & TAP\cite{zhou2018transferable}        & 63.0\% & 56.6\% &  50.4\%  & \bf{100.0}\%  & 92.8\%  \\
			& TAP*     & 77.6\%  & \bf{75.2}\% & 65.2\%    & \bf{100.0}\%  & \bf{98.8}\%  \\
			& ILA\cite{huang2019enhancing} & 72.3\%  & 69.5\% & 61.2\%    & \bf{100.0}\%  & 97.7\%  \\
			& DG-ILA(ours) & 74.1\%  & 71.7\% & 62.9\%    & \bf{100.0}\%  & 97.5\%  \\
			& ILA++\cite{li2020yet}  & 75.6\%  & 72.0\% & 65.1\%    & \bf{100.0}\%       & 98.3\%   \\
			& DG-ILA++(ours) & \bf{77.7}\%  & 74.5\%  & \bf{67.1}\% & \bf{100.0}\% & 98.5\%  \\
			\hline
			\hline
			\multirow{6}{*}{Res-152} & TAP\cite{zhou2018transferable}        & 63.9\% & 58.2\% &  53.6\%  & 94.9\%  & \bf{100.0}\%  \\
			& TAP*     & 72.1\%  & 70.9\% & 63.7\%    & 94.8\%  & 99.2\%  \\
			& ILA\cite{huang2019enhancing} & 73.7\%  & 73.8\% & 67.9\%    & 96.2\%  & 99.7\%  \\
			& DG-ILA(ours) & \bf{76.6}\%  & 77.4\% & 68.5\%    & 96.2\%  & 98.7\%  \\ 
			& ILA++\cite{li2020yet}  & 74.6\%  & 75.5\% & 68.0\%    & \bf{96.7}\%       & 99.8\%   \\
			& DG-ILA++(ours) & 76.5\%  & \bf{78.2}\%  & \bf{68.9}\% & 96.5\% & 99.2\%  \\ \bottomrule
		\end{tabular}%
	}
\end{wraptable}
In this subsection, we focus on evaluating our proposed DGM. Following the protocols in \cite{huang2019enhancing}, the same intermediate layer are selected for all the methods. Specifically, the selected intermediate layer for Inc-v3, Inc-v4, IncRes-v2, Res-101 and Res-152 are `Mixed6c', `feature-9', `mixed6a', `layer3' and `layer2', respectively. According to our experiments, the performance of constraining all the layers in original TAP \cite{zhou2018transferable} is actually lower than that of constraining certain layer alone, which is implemented by us and denoted as TAP*. Besides, the results of ILA \cite{huang2019enhancing} and ILA++ \cite{li2020yet} are reproduced with their released source codes.

\indent \textbf{Ablation Study.}
The results of ablation study are given in Table \ref{tab0}. As can be observed, both DG-ILA and DG-ILA++ obviously outperforms their original baseline methods. When the source model is IncRes-v2, the performance gain can reach as much as $5\%$. Besides, DG-ILA performs better than another related work TAP* in most cases and DG-ILA++ can further improve the performance. Apparently, our DGM can successfully boost the transferability of black-box attack methods.

\begin{figure}[tp]
	\centering
	\subfigure[\scriptsize Incv3->Incv4]{ 
		\includegraphics[width=0.23\linewidth,trim=0.2cm 0.2cm 2.6cm 2.0cm,clip]{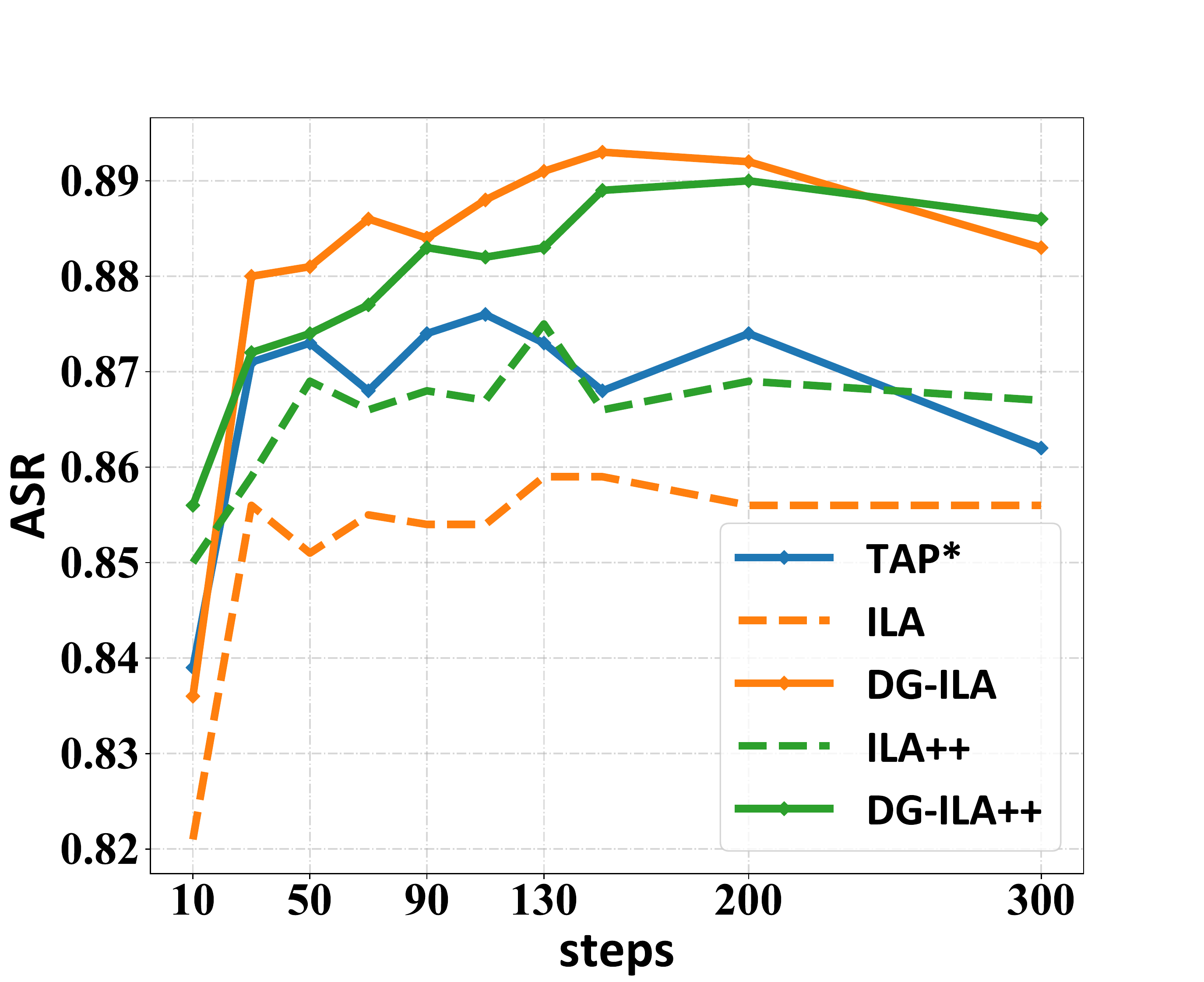}}
	\subfigure[\scriptsize Incv3->IncRes-v2]{
		\includegraphics[width=0.23\linewidth,trim=0.2cm 0.2cm 2.6cm 2.0cm,clip]{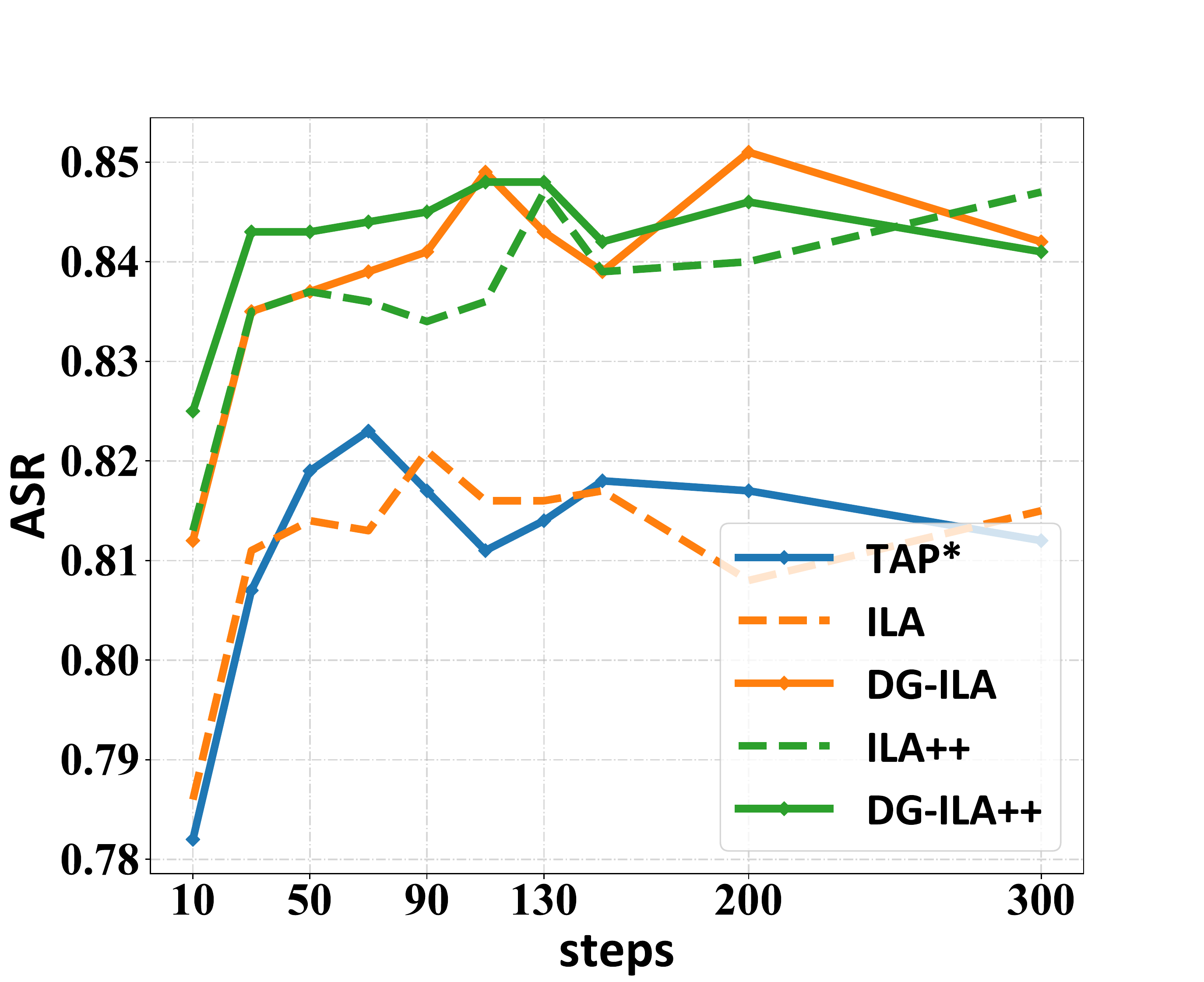}}
	\subfigure[\scriptsize Incv3->Res-101]{
		\includegraphics[width=0.23\linewidth,trim=0.2cm 0.2cm 2.6cm 2.0cm,clip]{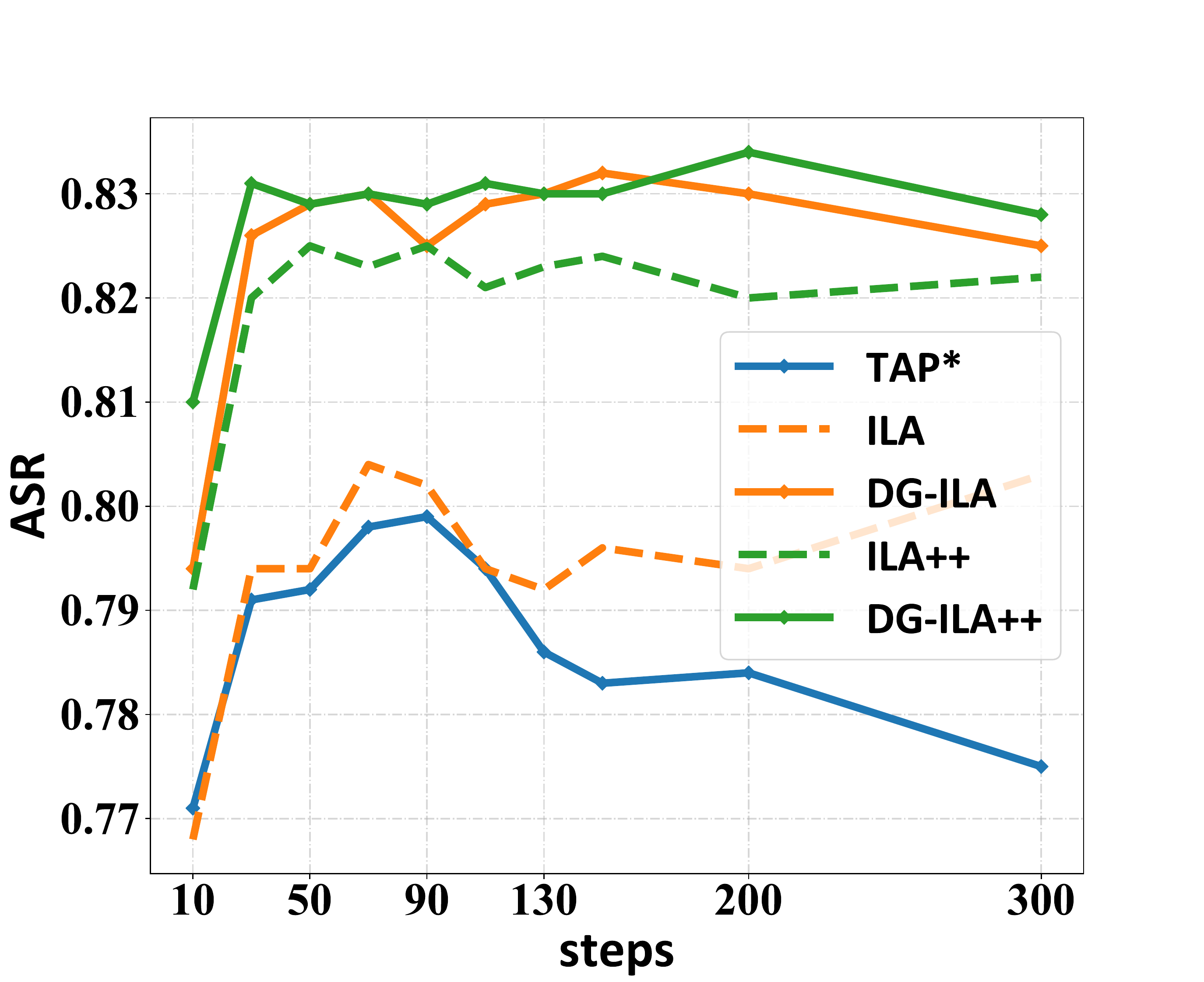}}
	\subfigure[\scriptsize Incv3->Res-152]{
		\includegraphics[width=0.23\linewidth,trim=0.2cm 0.2cm 2.6cm 2.0cm,clip]{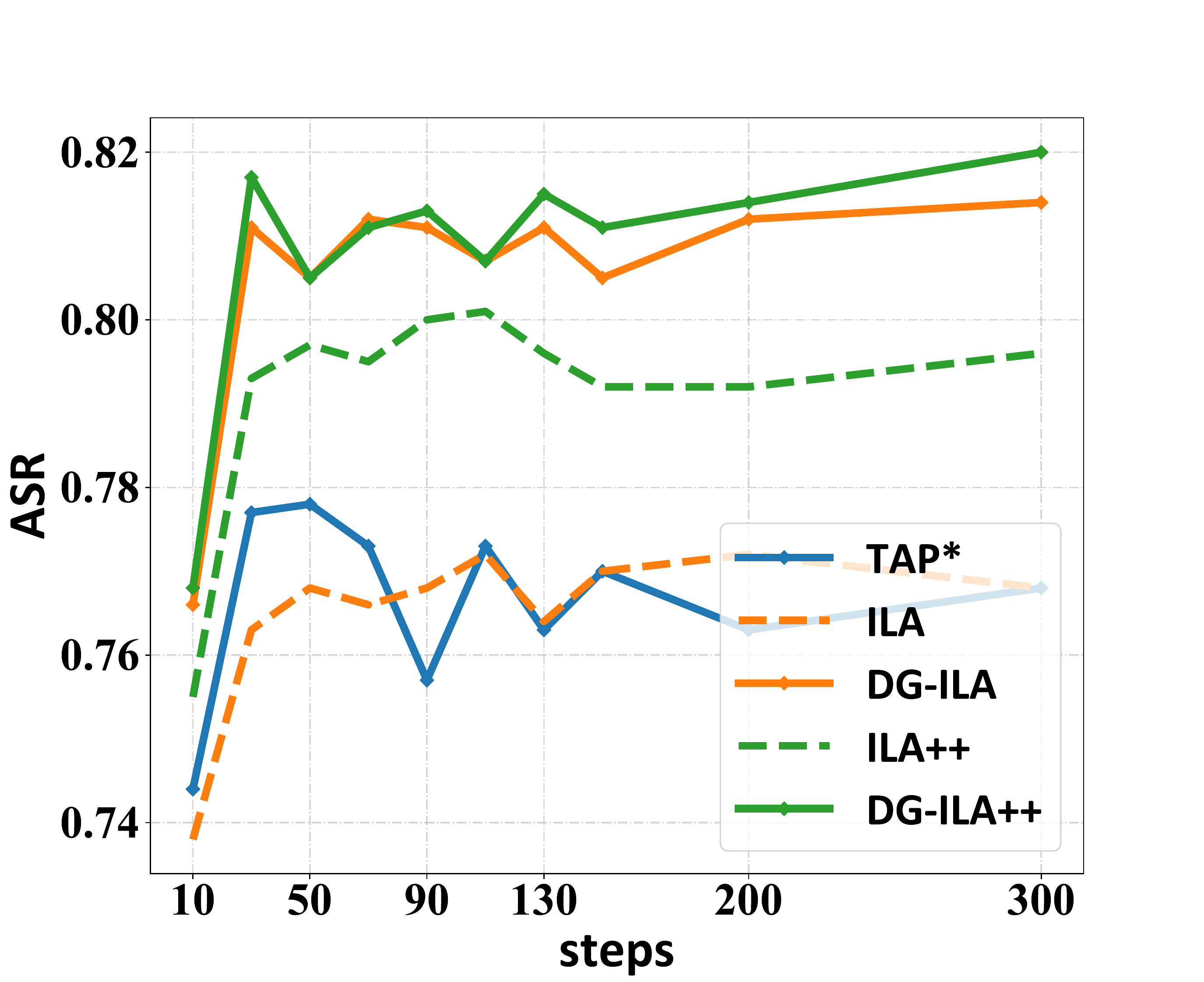}}
	\caption{Results of DGM when varying the number of iterations $T$. The caption `A'->`B' in each sub-figure indicates that `A' is the source model and `B' is the target model.}
	\label{Figure0} 
\end{figure}

\indent \textbf{Effects of the Number of Iterations.}
Figure \ref{Figure0} presents the steps-vs-ASR results. We can observe that DG-ILA has a sustainable advantage over ILA as the increase of iteration number. Similar tendency can also be seen from ILA++ and DG-ILA++. A possible explanation is that the current search direction, which is usually better than the guidance from baseline attack phase, will make contribution for next search and such progressive manner has more potential to find transferable directions. As for TAP*, it may fall into overfitting more easily because of fixed optimization objectives.

\begin{wrapfigure}[14]{r}{0.5\textwidth}
	\centering
	\subfigure[\scriptsize ]{ 
		\includegraphics[width=0.48\linewidth,trim=0.3cm 0.0cm 1.2cm 2.7cm,clip]{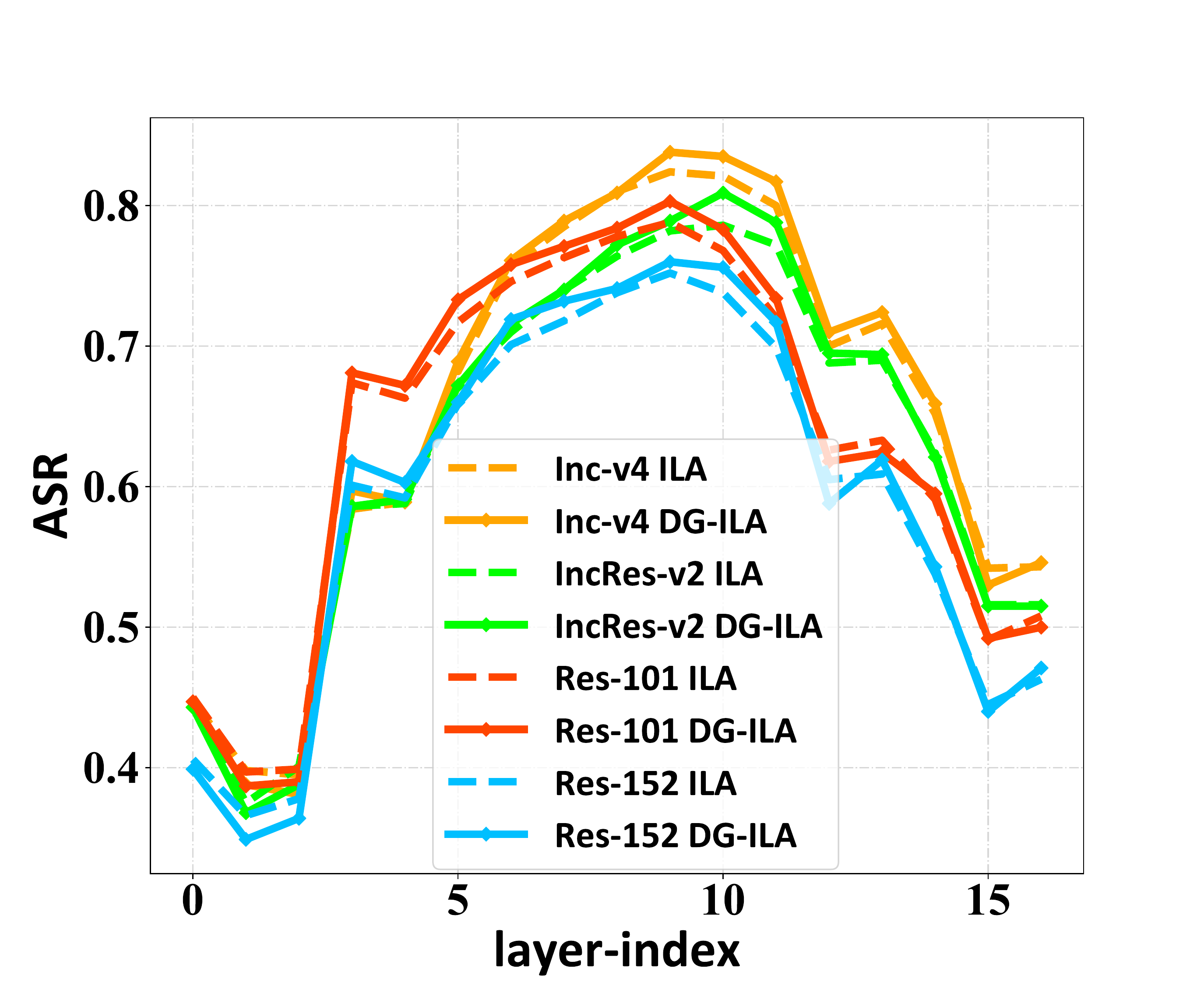}}
	\subfigure[\scriptsize ]{
		\includegraphics[width=0.48\linewidth,trim=0.3cm 0.0cm 1.2cm 2.7cm,clip]{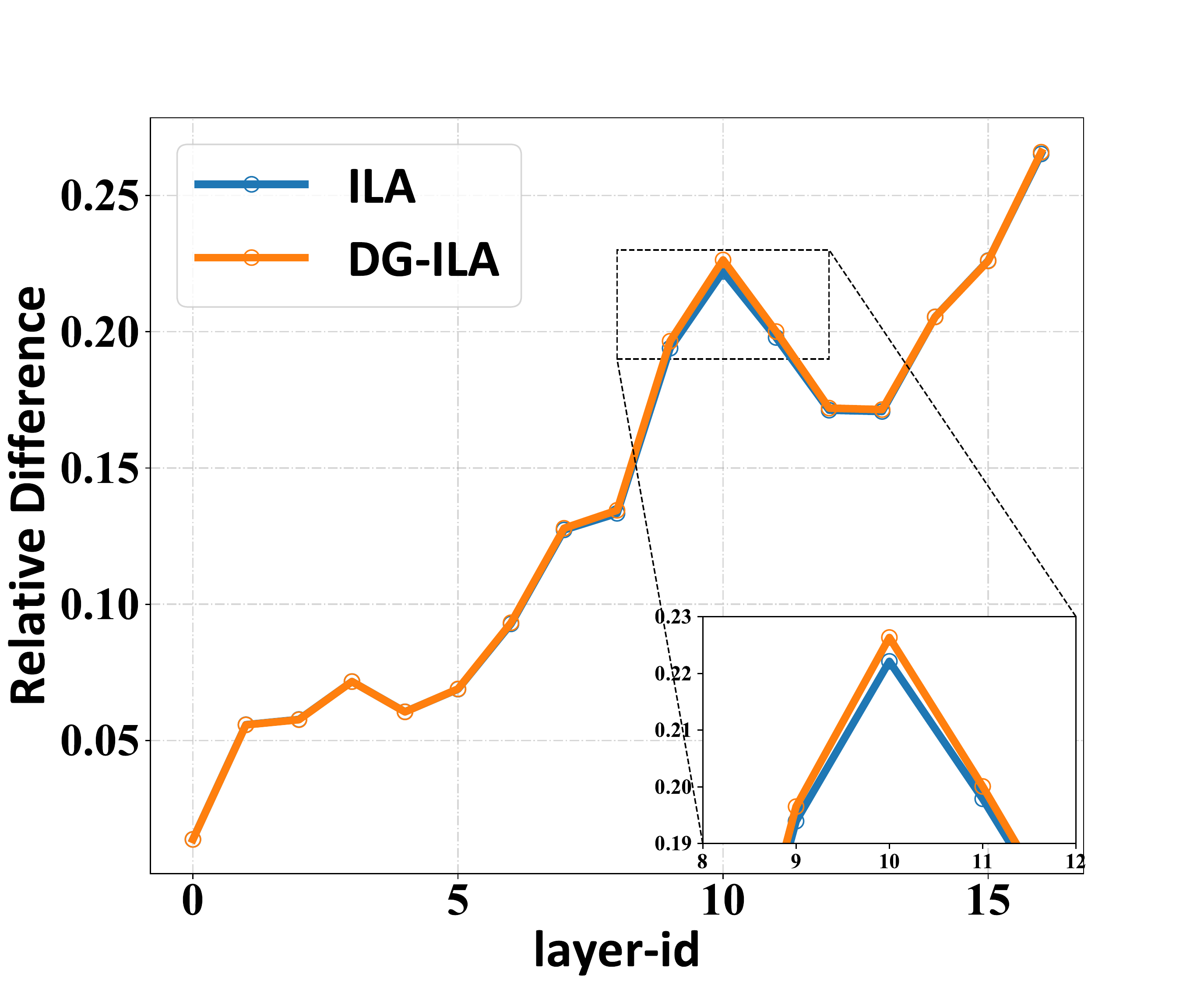}}
	\caption{(a) The attack success rate comparison across various layers. (b) The relative differences comparison of intermediate features.}
	\label{Fig_intermediate} 
\end{wrapfigure}

\indent \textbf{Effects of Different Layer Selections.}
Since ILA is a layer-centric attack method, the to-be-manipulated layer is specifically selected on the source model to obtain the best performance on the target model. Then, it is necessary to examine the performance of our DGM across various layers. Taking Inception-v3 as the source model, we select the layers from `Conv2d\_1a\_3x3' to `last\_linear', except for the pooling and dropout layers, as the intermediate layers for evaluation. The results are shown in Figure \ref{Fig_intermediate}(a). As can be observed, DG-ILA performs better than ILA in most of the selected layers and can give the best performance in black-box scenarios. Note that DG-ILA does not function very well at the bottom and top layers, which may be induced by the initial directional guidance is less transferable and lacks basic guidance ability. Fortunately, these layers will not be selected in practice because of their poor ASRs compared to the mid-level layers. To further verify the effectiveness of our DGM, we also calculate the relative feature difference, which is computed by
$\|\mathscr{F}^l(X^{adv}) - \mathscr{F}^l(X)\|_2/ \|\mathscr{F}^l(X)\|_2$, between the benign and the corresponding adversarial examples across various layers on Inception-v3. As can be observed from Figure \ref{Fig_intermediate}(b), DG-ILA gives larger mid-layer disturbance (in feature maps) than ILA, which is consistent with our expectation and the conclusion drawn in \cite{zhou2018transferable} and \cite{li2020yet}.

\begin{wraptable}[20]{r}{0.5\textwidth}
	\vspace{-0.9cm}
	\centering
	\caption{The evaluation of AIM and usefulness of single-form transformation adopted by us.}
	\label{tab1}
	\resizebox{0.5\textwidth}{!}{
		\begin{tabular}{@{}c|c|ccccc@{}}
			\toprule
			Model                   & Attack         & Inc-v3  & Inc-v4 & IncRes-v2 & Res-101 & Res-152 \\ \midrule
			\hline
			\multirow{6}{*}{Inc-v3} & MI-FGSM\cite{dong2018boosting}        & \bf{100.0\%} & 51.3\% & 49.2\%    & 46.8\%  & 42.4\%  \\
			& DI-MI-FGSM\cite{xie2019improving}     & 98.2\%  & 72.3\% & 69.5\%    & 59.2\%  & 58.1\%  \\
			& AI-MI-FGSM($A_1$) & 99.3\%  & 75.5\% & 72.5\%    & 61.9\%  & 60.6\%  \\
			& AI-MI-FGSM($A_2$) & 96.3\%  & 73.8\% & 72.0\%    & 63.3\%  & 62.4\%  \\
			& AI-MI-FGSM($A_3$) & 97.7\%  & 76.6\% & \bf{74.0\%}    & \bf{65.2\%}  & \bf{64.5\%}  \\
			& AI-MI-FGSM($A_4$) & 98.3\%  & \bf{76.8\%} & 73.3\%    & 63.7\%  & 64.3\%  \\
			\hline
			\hline
			\multirow{6}{*}{Inc-v4} & MI-FGSM\cite{dong2018boosting}        & 53.5\%  & \bf{100.0\%} & 47.3\%    & 45.9\%  & 41.8\%  \\
			& DI-MI-FGSM\cite{xie2019improving}     & 72.4\%  & 96.0\% & 69.1\%    & 59.2\%  & 59.1\%  \\
			& AI-MI-FGSM($A_1$) & \bf{76.9\%}  & 97.7\% & \bf{72.2\%}    & 63.2\%  & 61.4\%  \\
			& AI-MI-FGSM($A_2$) & 73.2\%  & 93.0\% & 69.9\%    & 63.7\%  & 60.8\%  \\
			& AI-MI-FGSM($A_3$) & 76.1\%  & 96.4\% & 70.9\%    & \bf{66.0\%}  & \bf{64.8\%}  \\
			& AI-MI-FGSM($A_4$) & 76.7\%  & 94.0\% & 70.9\%    & 64.3\%  & 62.5\%  \\
			\hline
			\hline
			\multirow{6}{*}{IncRes-v2} & MI-FGSM\cite{dong2018boosting}     & 58.6\% & 51.8\% & \bf{98.2\%}    & 46.6\%  & 45.8\%  \\
			& DI-MI-FGSM\cite{xie2019improving}     & 69.5\%  & 65.4\% & 87.6\%    & 55.7\%  & 55.9\%  \\
			& AI-MI-FGSM($A_1$) & \bf{73.9\%}  & \bf{72.4\%} & 91.2\%    & 61.3\%  & 62.0\%  \\
			& AI-MI-FGSM($A_2$) & 72.6\%  & 71.5\% & 88.0\%    & 63.4\%  & 63.6\%  \\
			& AI-MI-FGSM($A_3$) & \bf{73.9\%}  & 69.7\% & 88.8\%    & \bf{65.0\%}  & \bf{64.6\%}  \\
			& AI-MI-FGSM($A_4$) & 70.1\%  & 70.0\% & 87.4\%    & 61.3\%  & 62.0\%  \\
			\hline
			\hline
			\multirow{6}{*}{Res-101} & MI-FGSM\cite{dong2018boosting}       & 53.1\%   & 47.3\% & 39.3\%    & \bf{100.0\%}  & 90.5\%  \\
			& DI-MI-FGSM\cite{xie2019improving}     & 77.7\%  & 73.5\% & 67.8\%    & \bf{100.0\%}  & 94.7\%  \\
			& AI-MI-FGSM($A_1$) & 83.9\%  & \bf{82.4\%} & \bf{77.8\%}    & \bf{100.0\%}  & \bf{97.9\%}  \\
			& AI-MI-FGSM($A_2$) & 81.1\%  & 77.6\% & 70.5\%    & 99.3\%  & 93.3\%  \\
			& AI-MI-FGSM($A_3$) & 80.4\%  & 78.0\% & 74.1\%    & 99.5\%  & 94.6\%  \\
			& AI-MI-FGSM($A_4$) & \bf{82.2\%}  & 79.6\% & 74.8\%    & 99.7\%  & 94.1\%  \\
			\hline
			\hline
			\multirow{6}{*}{Res-152} & MI-FGSM\cite{dong2018boosting}       & 53.6\%   & 49.4\% & 41.9\%    & 92.1\%  & \bf{100.0\%}  \\
			& DI-MI-FGSM\cite{xie2019improving}     & 78.7\%  & 76.6\% & 71.4\%    & 96.1\%  & \bf{100.0\%}  \\
			& AI-MI-FGSM($A_1$) & \bf{84.4\%}  & \bf{82.6\%} & \bf{79.1\%}    & 96.7\%  & \bf{100.0\%}  \\
			& AI-MI-FGSM($A_2$) & 81.6\%  & 78.1\% & 74.2\%    & 94.9\%  & 99.3\%  \\
			& AI-MI-FGSM($A_3$) & 81.3\%  & 79.6\% & 73.2\%    & \bf{96.9\%}  & 99.7\%  \\
			& AI-MI-FGSM($A_4$) & 83.3\%  & 80.5\% & 76.9\%    & 95.2\%  & 99.8\%  \\ \bottomrule
		\end{tabular}%
	}
\end{wraptable}

\subsection{Evaluation of AIM}

\indent \textbf{Ablation Study.}
We compare three attack methods, MI-FGSM \cite{dong2018boosting}, DI-MI-FGSM (momentum version of DIM) \cite{xie2019improving} and our AI-MI-FGSM, and present quantitative results in Table \ref{tab1}. Note that $A_i$ denotes the employed single-form transformation $A_i$ defined in \ref{eq1} to \ref{eq4}. From Table \ref{tab1}, a first glance shows that DI-MI-FGSM outperforms MI-FGSM by a large margin and AI-MI-FGSM is comparable to DI-MI-FGSM and even better than it in most black-box settings. This table verifies the usefulness of our each single-form operation when boosting the transferability of adversarial examples.

\begin{figure}[tp]
	\centering
	\subfigure[\scriptsize Incv3->Incv4]{ 
		\includegraphics[width=0.23\linewidth,trim=0.8cm 0.3cm 2.6cm 2.6cm,clip]{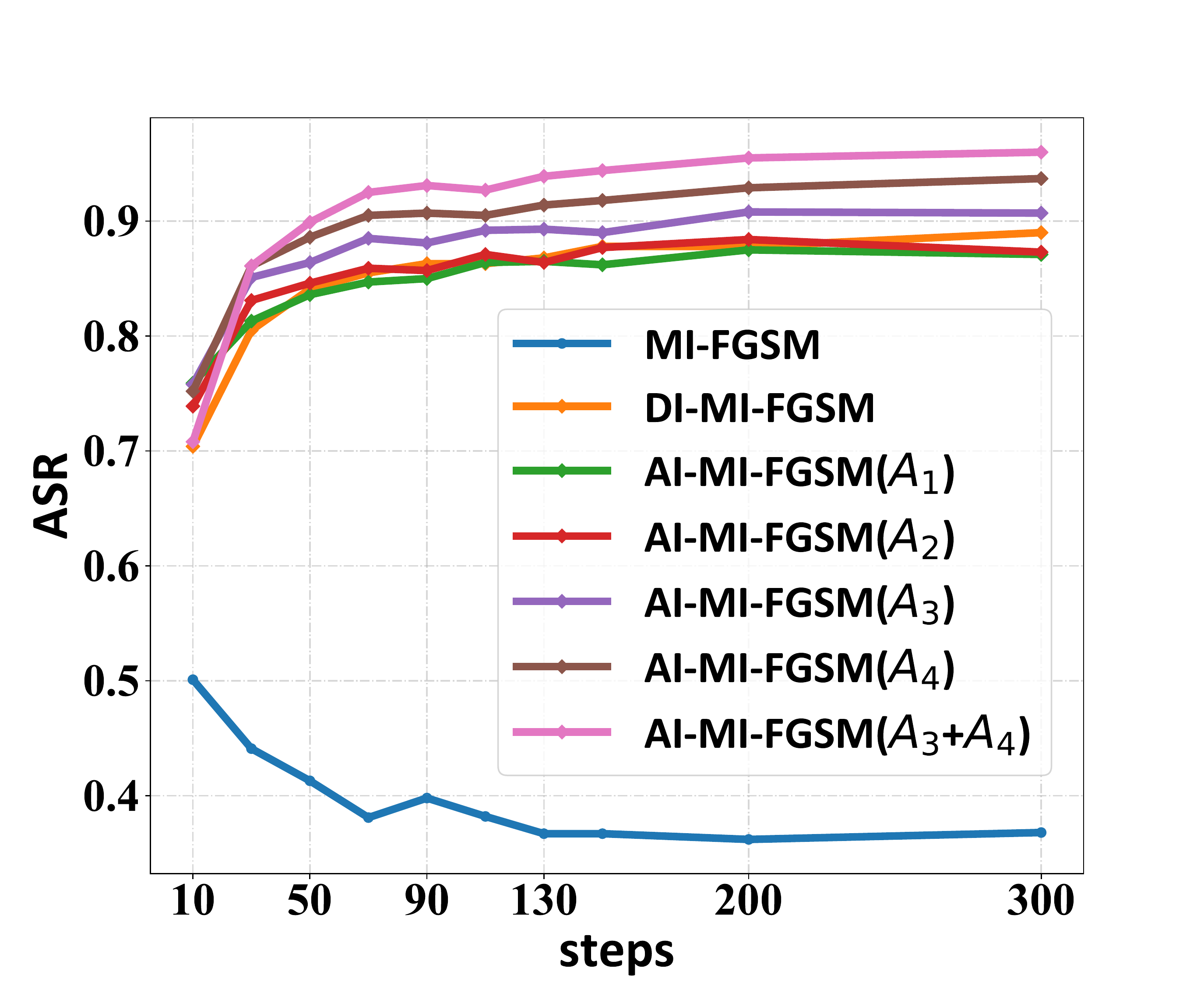}}
	\subfigure[\scriptsize Incv3->IncRes-v2]{
		\includegraphics[width=0.23\linewidth,trim=0.8cm 0.3cm 2.6cm 2.6cm,clip]{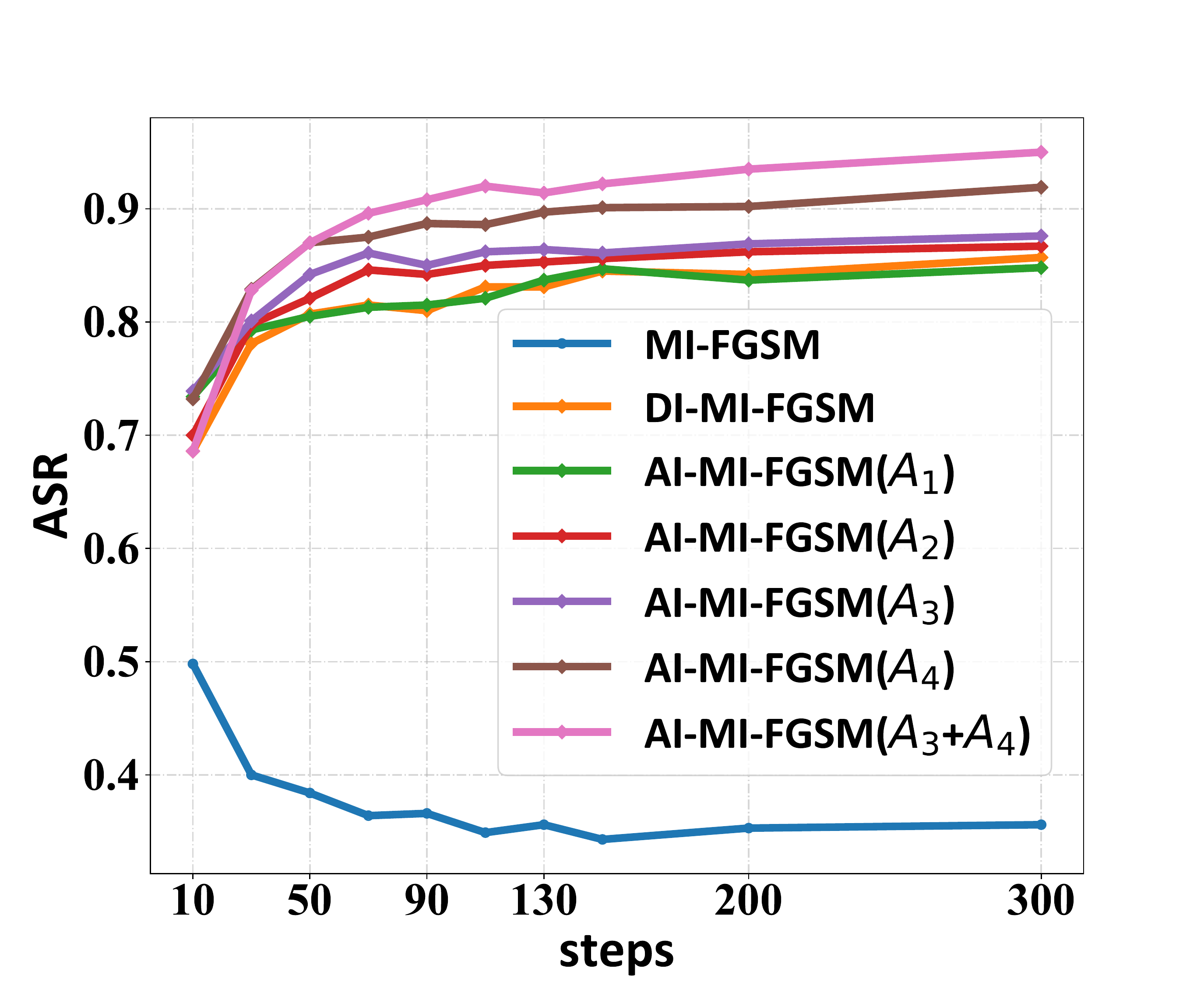}}
	\subfigure[\scriptsize Incv3->Res-101]{
		\includegraphics[width=0.23\linewidth,trim=0.8cm 0.3cm 2.6cm 2.6cm,clip]{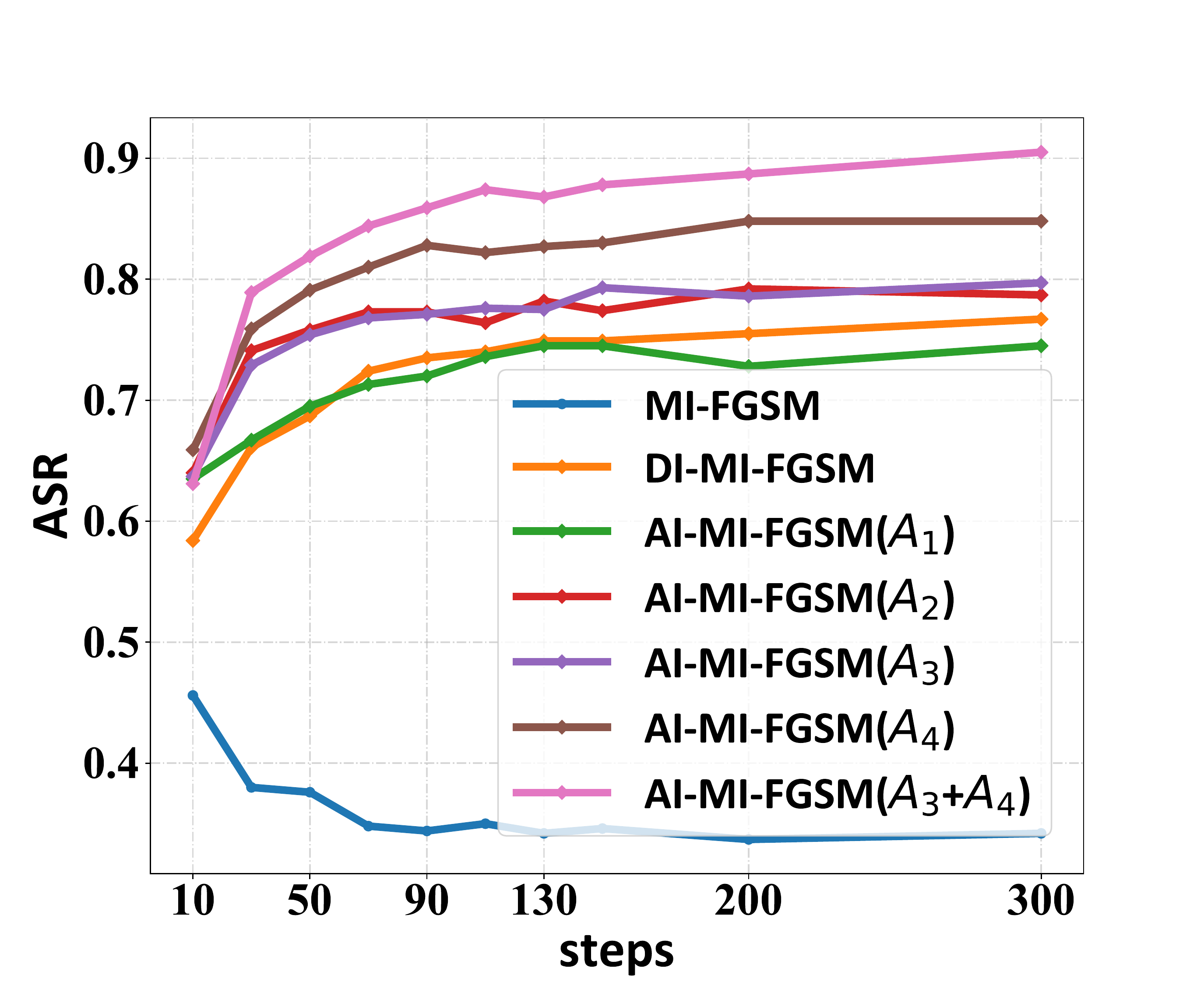}}
	\subfigure[\scriptsize Incv3->Res-152]{
		\includegraphics[width=0.23\linewidth,trim=0.8cm 0.3cm 2.6cm 2.6cm,clip]{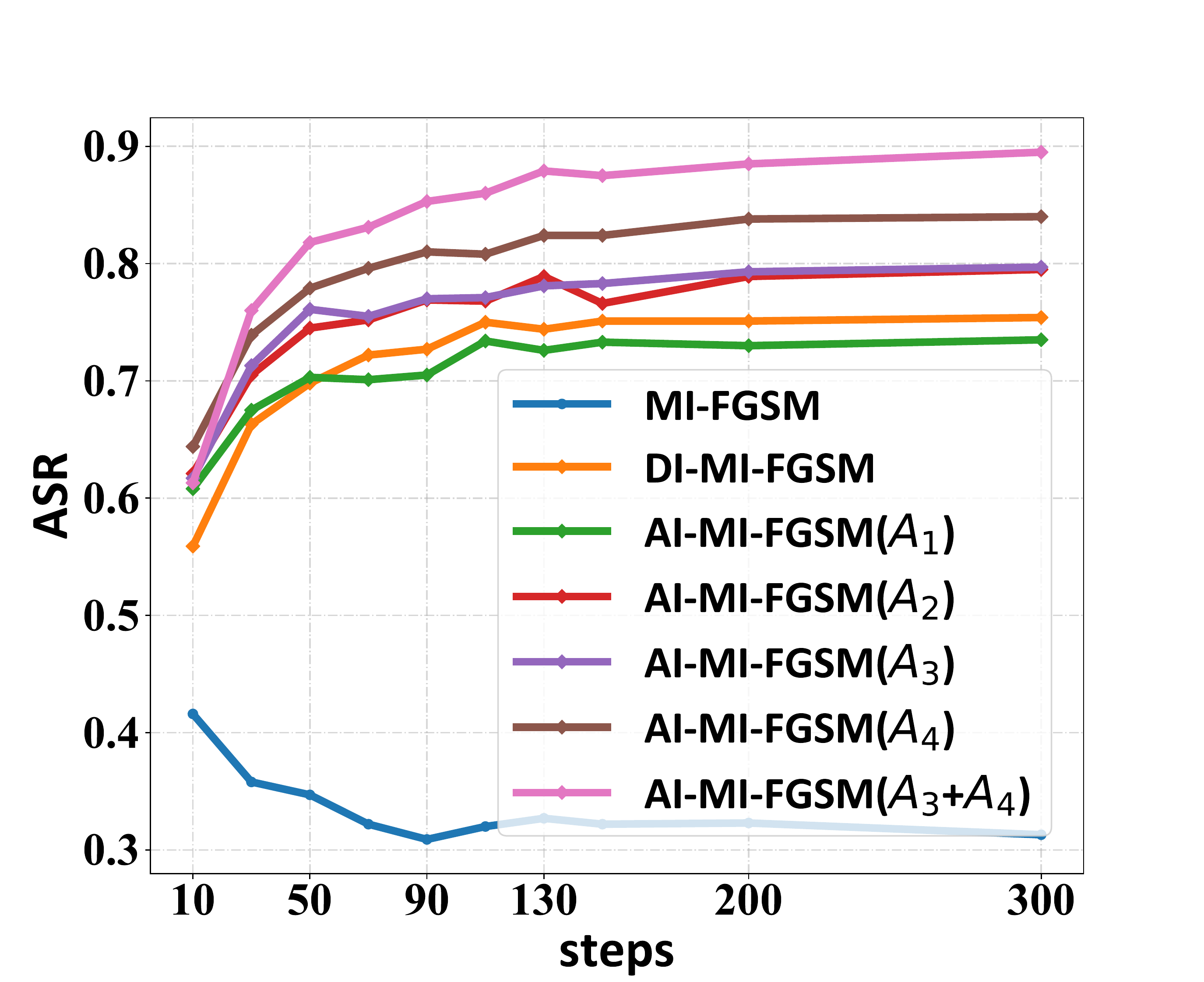}}
	\caption{Evaluation results of AIM and superiority of multi-form transformation when varying the iteration number $T$. The sub-script `A'->`B' below each sub-plot represents that A is the source model and B is the target model.}
	\label{Figure1} 
\end{figure}

\indent \textbf{Multi-form Operation Analysis.}
To further validate the effectiveness of the multi-form affine transformation augmentation, Inception-v3 is selected and the number of iterations are varied. The results are presented in Figure \ref{Figure1}. As can be observed, if the multi-form transformation is utilized, e.g., 'scaling+shearing', the attack success rates are superior for most of the iterations. Note that the performance of the multi-form transformation at the beginning is lower than the single-form transformations, which may be induced by the augmented input space with more diversity. Besides, MI-FGSM \cite{dong2018boosting} tends to fall into the overfitting dilemma when the number of iterations become large, while DI-MI-FGSM \cite{xie2019improving} can slightly alleviate this problem. In general, it is undeniable that AI-MI-FGSM with multi-form transformation surpasses DI-MI-FGSM \cite{xie2019improving} and further boost the transferability.

\begin{wrapfigure}[16]{r}{0.5\textwidth}
	\centering
	\subfigure[\scriptsize ]{ 
		\includegraphics[width=0.48\linewidth,trim=0.3cm 0.2cm 2.6cm 2.7cm,clip]{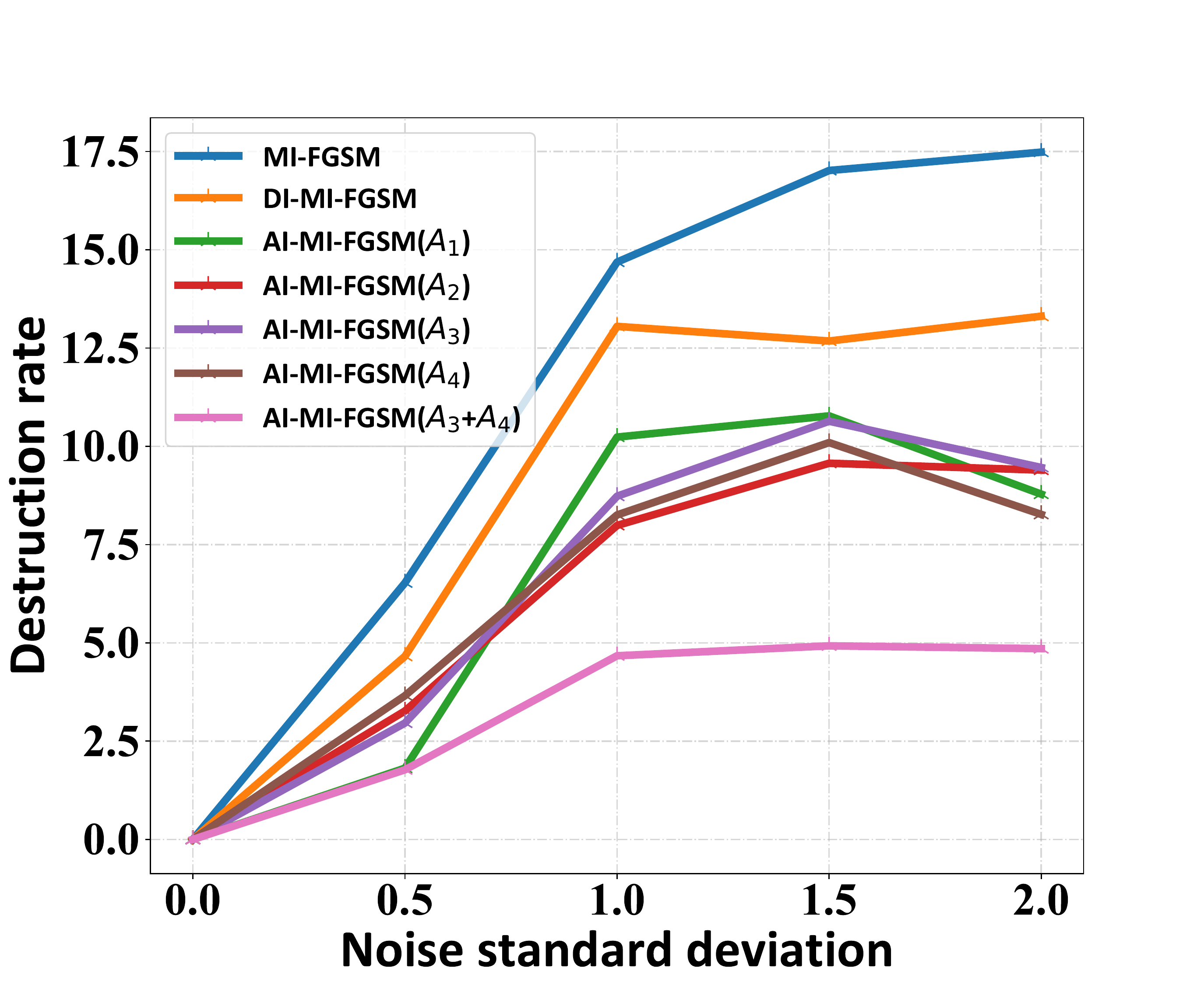}}
	\subfigure[\scriptsize ]{
		\includegraphics[width=0.48\linewidth,trim=0.3cm 0.2cm 2.6cm 2.7cm,clip]{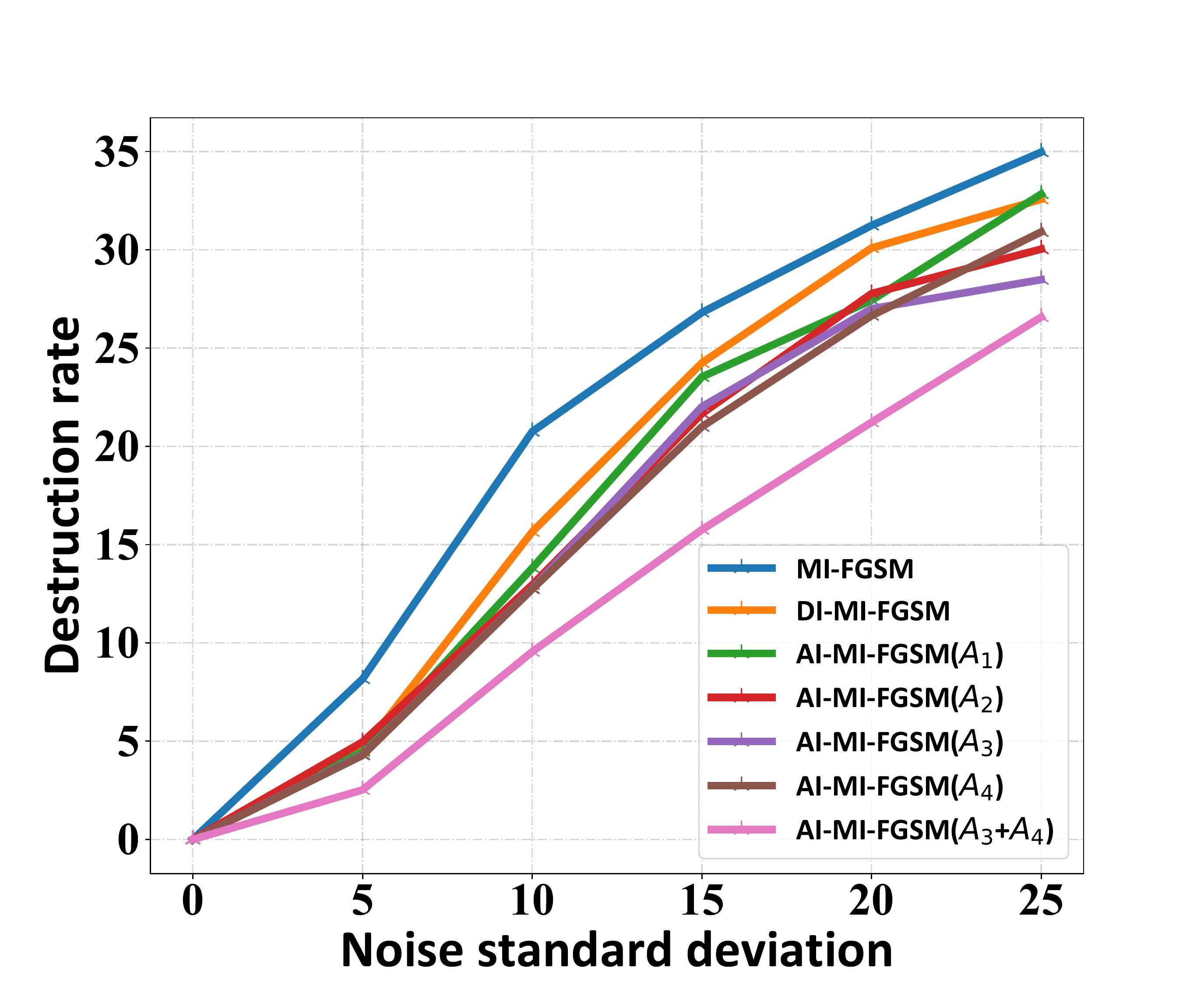}}
	\caption{Comparison of destruction rate for various methods under corruption. The lower the curve, the more robust against Gaussian corruption.}
	\label{Figure2} 
\end{wrapfigure}
\indent \textbf{Robustness Analysis.}
To evaluate the robustness of the generated adversarial examples from our method, we adopt Gaussian noise and Gaussian blurring to corrupt the generated adversarial examples to observe their attacking performances. Here, we uses the destruction rate, which is proposed in \cite{kurakin2016adversarial,wu2018understanding}, to quantify the robustness. Without loss of generality, we select Inception-v3 and Res-101 as the source and target models respectively. $1000$ images, which is the same as previous experiments, are still utilized. The destruction results under Gaussian blurring and Gaussian noise are shown in Figure \ref{Figure2}(a) and Figure \ref{Figure2}(b) respectively. As can be observed, the adversarial examples synthesized by AI-MI-FGSM are more robust than the baseline methods. Note that our multi-form transformation, e.g. 'scaling+shearing', obviously boosts the performance.


\subsection{Evaluation of TRAP}

\begin{figure*}[tb]
	\centering
	\subfigure[\scriptsize Incv3->Incv4]{ 
		\includegraphics[width=0.23\linewidth,trim=0.3cm 0.3cm 2.6cm 2.6cm,clip]{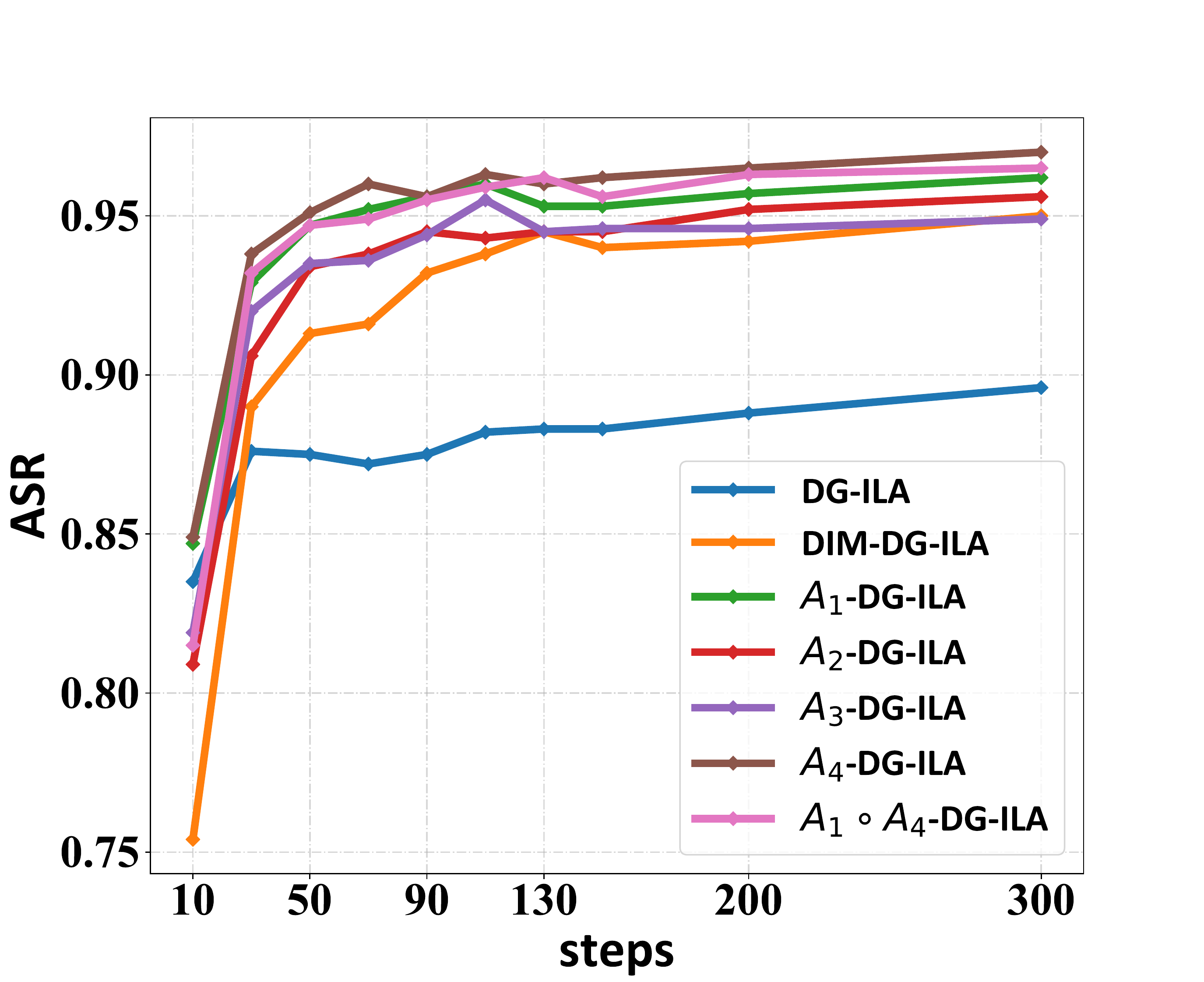}}
	\subfigure[\scriptsize Incv3->IncRes-v2]{
		\includegraphics[width=0.23\linewidth,trim=0.3cm 0.3cm 2.6cm 2.6cm,clip]{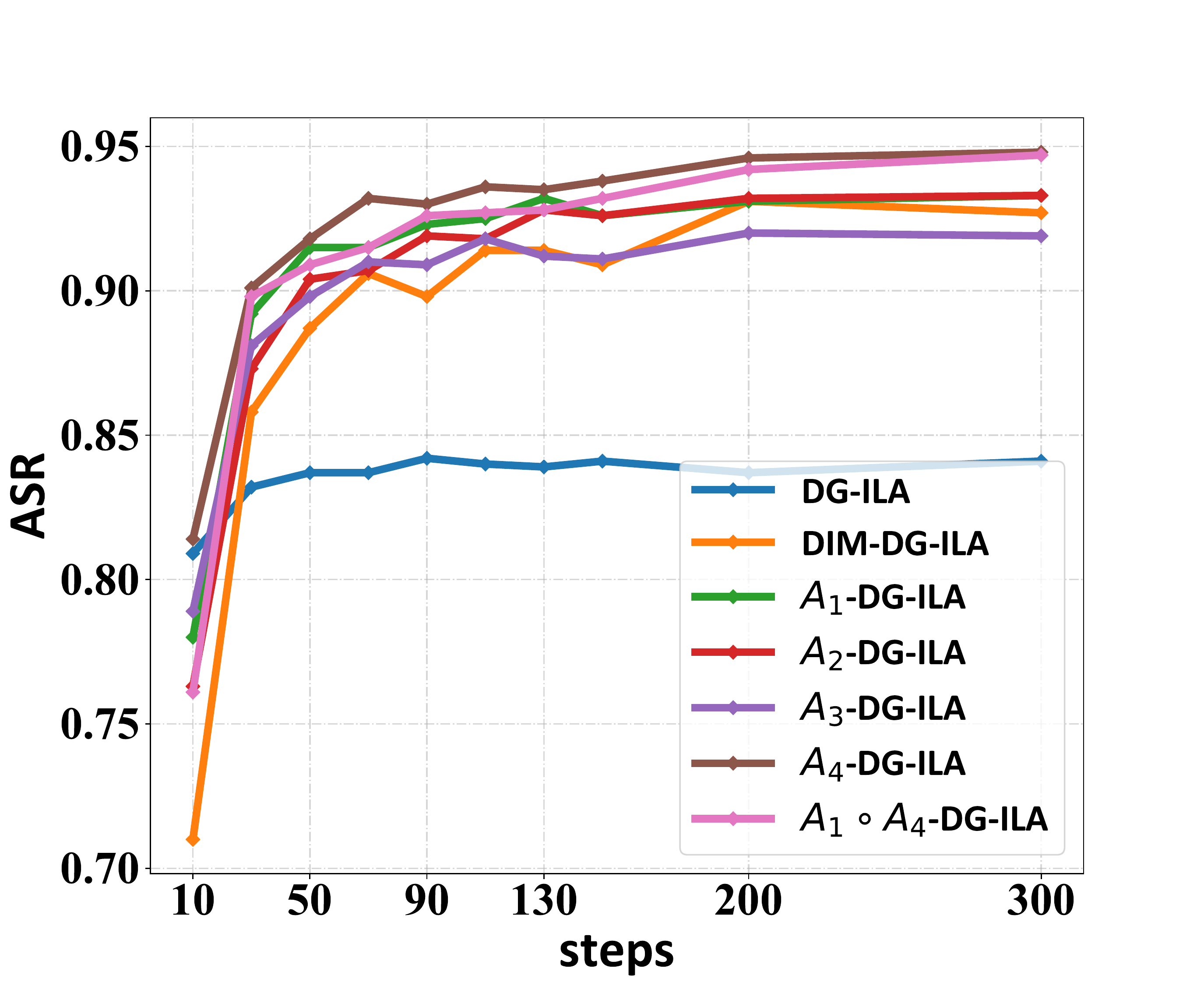}}
	\subfigure[\scriptsize Incv3->Res-101]{
		\includegraphics[width=0.23\linewidth,trim=0.3cm 0.3cm 2.6cm 2.6cm,clip]{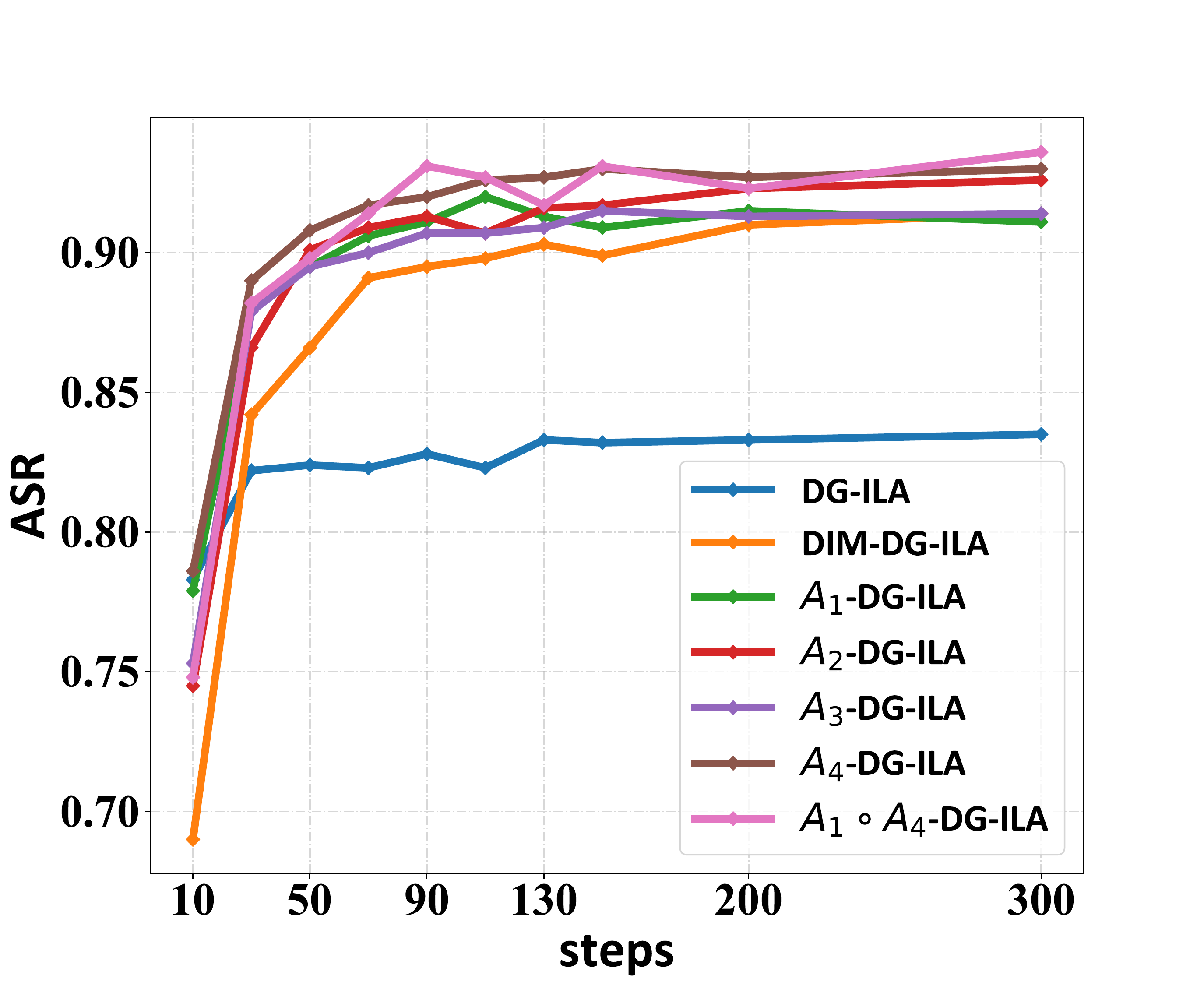}}
	\subfigure[\scriptsize Incv3->Res-152]{
		\includegraphics[width=0.23\linewidth,trim=0.3cm 0.3cm 2.6cm 2.6cm,clip]{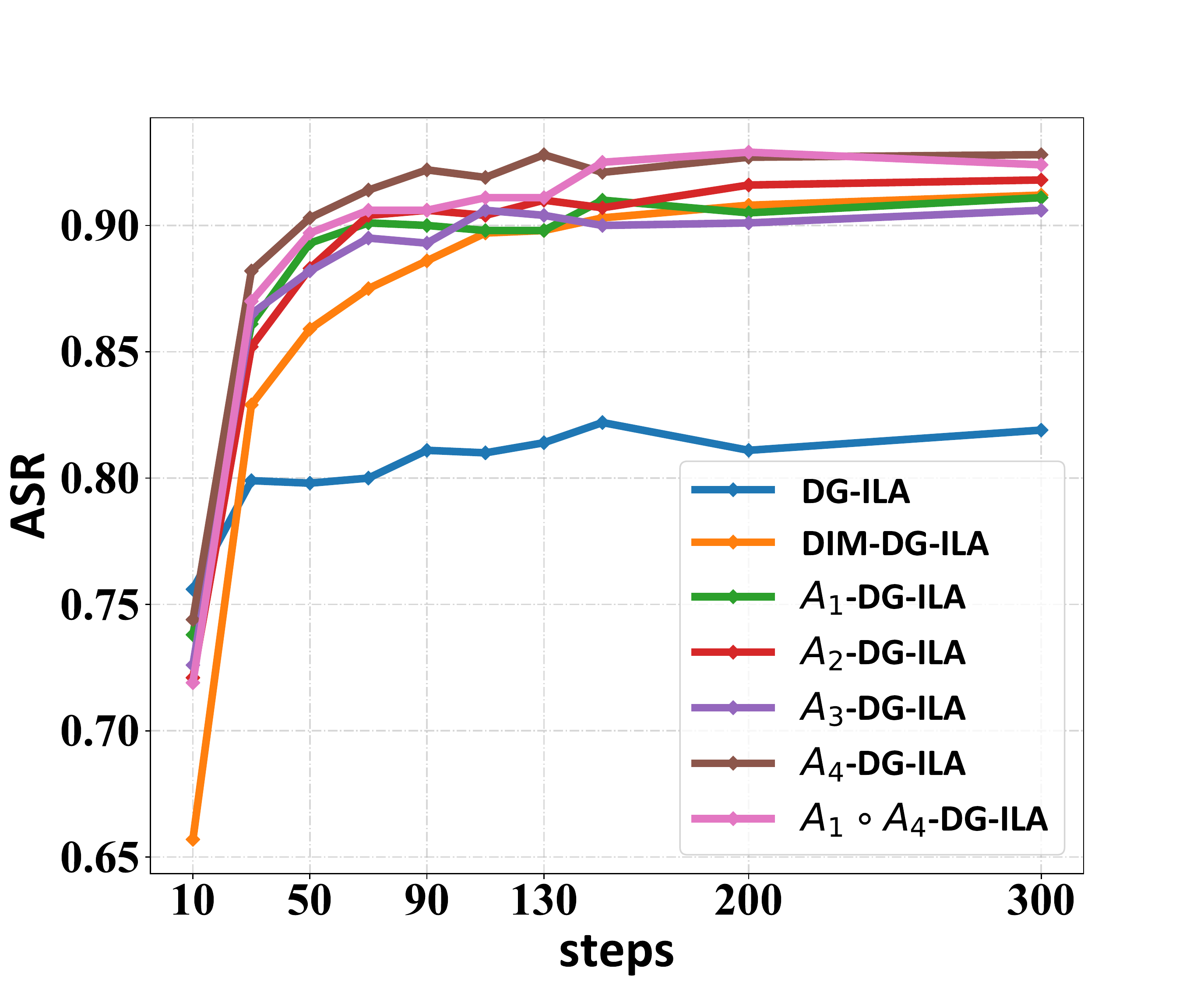}}
	\caption{Combination results of DG-ILA and various input transformation based methods. The caption `A'->`B' in each sub-figure represents that `A' is the source model and `B' is the target model.}
	\label{Figure3} 
\end{figure*}

\indent \textbf{Effects of the Combination of DGM and AIM.}
To verify the effectiveness of combining the two proposed mechanisms, i.e., DGM and AIM, we select different input transformations and execute Algorithm \ref{algorithm1} to generate the final adversarial examples. The results are shown in Figure \ref{Figure3}. As can be observed, the obvious performance improvement indicates that our two mechanisms can collaborate complementarily, even with single basic transformations. If we exploit the multi-form affine transformation in the combination, the results are comparative with the best performance of single-form and even better, especially on Res-101.

\indent \textbf{Comparison with State-of-the-Arts.}
Here, the proposed TRAP is compared to the existing SOTAs on five models, which are employed in the previous experiments, as well as three defense models, i.e., adv-ResNet152 Baseline (adv-Res152B), adv-ResNet152 Denoise (adv-Res152D), adv-ResNeXt101 DenoiseAll (adv-ResNeXtDA) \cite{xie2019feature}. Note that we have subtracted the ratio of wrongly predicted benign images in the reported ASR results. Because these models are very robust and in order to reach full convergence, we perform 300 steps for all comparative attack methods. As can be observed from Table \ref{tab2}, our TRAP significantly outperforms the existing SOTAs on the normally trained models in most of the cases. Meanwhile, our TRAP gives better performance on the defended models in most of the cases.

\begin{table*}[tb]
	\centering
	\caption{The ASRs of our TRAP and other SOTAs on five normally trained models and three defense models.}
	\label{tab2}
	\resizebox{0.95\textwidth}{!}{
		\begin{tabular}{@{}c|c|cccccccc@{}}
			\toprule
			Model                   & Attack         & Inc-v3  & Inc-v4 & IncRes-v2 & Res-101 & Res-152 & adv-Res152B & adv-Res152D & adv-ResNeXtDA \\ \midrule
			\hline
			\multirow{7}{*}{Inc-v3} & MI-FGSM\cite{dong2018boosting}        & \bf{100.0\%} & 36.8\% & 35.6\%    & 34.2\%  & 31.3\%  &0.8\% & 0.7\% & 0.7\% \\
			& DI-MI-FGSM\cite{xie2019improving}     & \bf{100.0\%}  & 89.0\% & 85.7\%    & 76.7\%  & 75.4\%  & 2.7\% & 2.7\% & 2.9\% \\
			& TI-MI-FGSM\cite{dong2019evading}     & \bf{100.0\%}  & 40.0\% & 36.9\%    & 35.9\%  & 33.9\%  & 0.4\% & 0.8\% & 1.1\% \\
			& TAP\cite{zhou2018transferable}        & 99.7\%  & 86.2\%      & 81.2\%         & 77.5\%       & 76.8\%      & 1.9\%     & 2.4\%  &    2.1\%     \\
			& ILA\cite{huang2019enhancing}                            & 99.8\%   & 85.6\% & 81.5\%    & 80.3\%  & 76.8\%  & 2.3\% & 2.5\% & 2.4\% \\
			& ILA++\cite{li2020yet}                          & 99.8\%   & 86.7\% & 84.7\%    & 82.2\%  & 79.6\%  & 2.5\% & 2.8\% & 2.7\% \\
			& TRAP(ours)                        & 99.7\%   & \bf{96.5}\% & \bf{94.7\%}    & \bf{93.6\%}  & \bf{92.4\%}  & \bf{3.7\%} & \bf{4.6\%} & \bf{3.1\%} \\ 
			\hline 
			\hline
			\multirow{7}{*}{Res-101} & MI-FGSM\cite{dong2018boosting}       & 44.2\%   & 37.6\% & 31.3\%     & \bf{100.0}\%  & 86.3\%  & 0.9\% & 1.1\% & 1.3\%  \\
			& DI-MI-FGSM\cite{xie2019improving}     & \bf{93.5}\%  & 93.6\% & 88.7\%    & \bf{100.0}\%  & \bf{100.0}\% & 2.6\% & 3.1\% & \bf{4.4}\% \\
			& TI-MI-FGSM\cite{dong2019evading}     &  52.3\% & 46.0\% & 40.2\%    &\bf{100.0}\%  & 88.8\%  & 1.4\% & 1.7\% & 2.8\% \\
			& TAP\cite{zhou2018transferable}                            &  76.3\% & 74.9\%      &  62.1\%  & \bf{100.0}\%  & 98.9\%    &0.7\%   &1.6\%  & 1.6\%  \\
			& ILA\cite{huang2019enhancing}                            & 75.1\%   & 73.1\% & 65.3\%    & \bf{100.0}\%  & 98.7\%  & 1.1\% & 1.7\% & 2.4\% \\
			& ILA++\cite{li2020yet}                          & 77.5\%   & 75.5\% & 68.1\%    & \bf{100.0}\%  & 99.2\%  & 1.3\% & 1.7\% & 2.7\% \\ 
			& TRAP(ours)                                           & 93.0\% & \bf{93.8}\% & \bf{89.6}\% & \bf{100.0}\% & \bf{100.0}\% & \bf{3.8}\% & \bf{3.5}\%  & 3.4\%  \\         \bottomrule
		\end{tabular}%
	}
\end{table*}

\section{Conclusion}
\label{conclusion}
In this work, we investigate to improve the transferability and robustness of adversarial examples from the perspective of network hierarchy. In the intermediate stage of network, we propose a dynamically guided mechanism to iteratively revising the directional guidance during the perturbation generation process. In the input stage of network, we propose to adopt multi-form affine transformation to augment the input images to enrich the input diversity. Experimental results have demonstrated the effectiveness of our proposed DGM and AIM, as well as the transferability and robustness of our proposed TRAP.
{\small
	\bibliographystyle{ieee}
	\bibliography{arxiv2021}
}


\end{document}